\documentclass[final,5p,times,twocolumn]{elsarticle}

\usepackage{multirow}
\usepackage{amsmath}
\usepackage{graphicx}
\usepackage{natbib}
\usepackage{marvosym}
\usepackage{amssymb}
\usepackage{booktabs} 
\usepackage{multirow}
\usepackage{color}
\usepackage{hyperref}

\usepackage[linesnumbered,ruled]{algorithm2e}
\SetKwProg{Fn}{Function}{}{end}\SetKwFunction{FRecurs}{FnRecursive}%

\usepackage{etoolbox}
\usepackage{flushend}

\journal{XXX}

\begin{document}


\begin{frontmatter}



\title{Modeling and Multi-objective Optimization of a Kind of Teaching Manipulator}







\author[addr1]{Zhun Fan\corref{cor1}}
\ead{zfan@stu.edu.cn}

\author[addr1]{Yugen You}
\author[addr1]{Haodong Zheng}
\author[addr1]{Guijie Zhu}
\author[addr1]{Wenji Li}
\author[addr1]{Shen Chen}
\author[addr2]{Kalyanmoy Deb}
\author[addr2]{Erik Goodman}

\cortext[cor1]{Corresponding author}

\address[addr1]{Department of Electronic Engineering, Shantou University, Guangdong, China}

\address[addr2]{BEACON Center for the Study of Evolution in Action, Michigan State University. East Lansing, Michigan, USA.}

\begin{abstract}
A new kind of six degree-of-freedom teaching manipulator without actuators is designed, for recording and conveniently setting a trajectory of an industrial robot. The device requires good gravity balance and operating force performance to ensure being controlled easily and fluently. In this paper, we propose a process for modeling the manipulator and then the model is used to formulate a multi-objective optimization problem to optimize the design of the testing manipulator. Three objectives, including total mass of the device, gravity balancing and operating force performance are analyzed and defined. A popular non-dominated sorting genetic algorithm (NSGA-II-CDP) is used to solve the optimization problem. The obtained solutions all outperform the design of a human expert. To extract design knowledge, an innovization study is performed to establish meaningful implicit relationship between the objective space and the decision space, which can be reused by the designer in future design process.
\end{abstract}


\end{frontmatter}


\section{Introduction}
\label{sec:intro}
\begin{figure*}
\centering
\includegraphics [width=15cm]{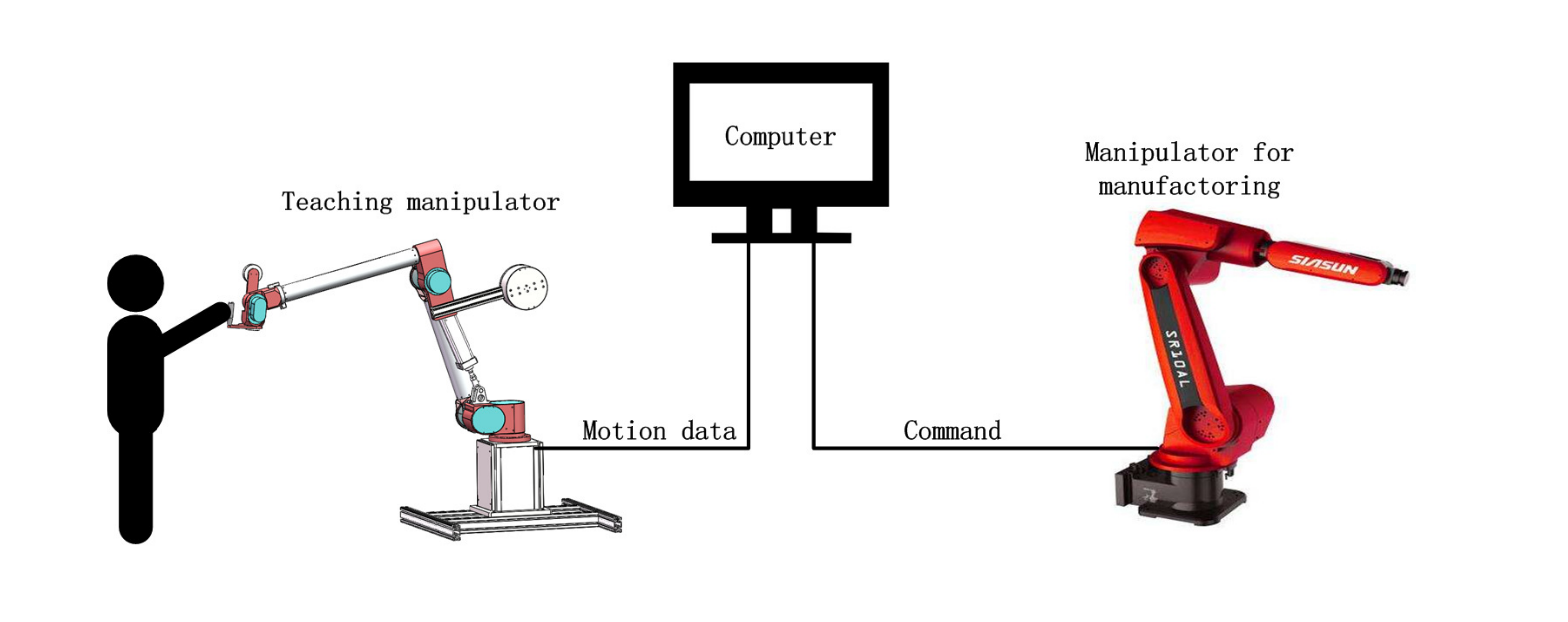}
\caption{The operator controls the teaching manipulator to record the trajectory. The motion data is saved in the computer and used for command the real manipulator for manufacturing.}
 \label{work}
\end{figure*}

For manufacturers who run for customized production, designing and setting manipulator trajectories in the programming system is a tedious and time-consuming task, because trajectories need to be redesigned and reset frequently to adapt to different applications in a flexible production line. To simplify the process, a teaching manipulator with six-degree-of-freedom (6-DOF) is designed in this paper. The device is designed for recording trajectories and teaching a real robot to accomplish the task, as shown in Fig. \ref{work}. With its help, operators can conveniently set a proper trajectory through conducting the teaching manipulator to finish the task for the first time. Then a real robot or a batch of robots will follow the speed and the trajectory of the teaching robot to accomplish the task for many other products. A teaching manipulator with lighter weight and better operating performance is always required. However, the human expert cannot ensure the best performance in his or her design. Robot design automation is a systematic process of design optimization that can help to achieve a better design of the teaching manipulator.

Robot design automation is an emerging technology involving systematic modeling and optimization efforts. Kinematics, dynamics and stiffness are usually considered in modeling robots or other mechanical systems. A lot of progress has been made in this direction. For example, Pettersson et al \cite{pettersson2009drive} built up a model of the drive chain of light weight robotic arm for optimizing is design. Citalan-Lara et al \cite{citalan2014multidisciplinary} proposed analytical modeling of the mechanism, the controller and the servo drive subsystem of a kind of manipulator and optimized the manipulator with six objectives simultaneously. The research of parallel robot modeling and optimization has also made progress. Qin et al \cite{qin2013modelling} proposed a two-staged model for parallel mechanism with a rigid and a compliant platform. Laski et al \cite{laski2015design} designed and analyzed a kind of 3-DOF tripod parallel manipulator. Yao et al \cite{yao2017dynamic} established the dynamic driving force model of the a parallel manipulator with redundant actuation. To fully integrate advantages of both the serial and parallel robot structures, hybrid robots or robotic machine was developed. Gao et al \cite{gao2010design, gao2015performance} made a detailed analysis of a hybrid robotic machine tool and optimized its dimensional parameters.

Design optimization is an essential sub-process in design automation. Engineering solutions are expected to achieve good performance in a number of aspects, while satisfying various constraints at the same time. Therefore, multi-objective optimization algorithms are widely used to search for a group of non-dominated solutions, namely, Pareto-optimal front, considering the tradeoff among multiple objectives at the same time. Different algorithms is designed for solving robot optimization problem. Coello et al \cite{coello1998using} used a new genetic algorithm(GA)-based multi-objective optimization technique to optimize the counterweight balancing for a serial robot. Gao et al \cite{gao2010design} conducted an optimization study of system stiffness and dexterity for the parallel mechanism using multi-objective optimization. Zhang et al \cite{zhang2012forward} used a multi-objective optimization algorithm for optimizing a bio-inspired parallel manipulator design. Li and Xu \cite{li2006ga} proposed a GA-based multi-objective optimization approach to optimize a kind of cable-driven parallel manipulator. jamwal et al \cite{jamwal2015three} used NSGA-II to optimize a kind of rehabilitation robot considering six different performances and the botained solutions is better than the results obtained from sigle objective optimization and preference-based  optimization.

The multi-objective optimization is not only for providing design candidates of the problem, but also for providing design principles among optimal trade-off solutions. Innovization study is first proposed by Deb and Srinivasan \cite{deb2006innovization}, with the purpose to establish meaningful knowledge between objective functions and design variables. The knowledge can provide a deeper understanding about how the variables of the optimal solutions interact, which can help the designer acquire design insights. A robot gripper optimization problem \cite{osyczka2002evolutionary} is used as an example to conduct research on innovization study by Datta et al \cite{datta2011multi}. A further research performs the innovization study on the modified gripper model considering different actuators models \cite{datta2016piezoelectric, datta2016analysis}. Besides the gripper model, Deb et al \cite{deb2014integrated} also studied innovization with three other engineering problems.

As a case study, we perform modeling, structure analysis and optimization for a teaching manipulator with six degrees of freedom. Three performances, including total mass of the device, the maximal value of operating force and the difference between the maximum and minimum of operating force are treated as the objectives, constrained by the condition of gravity balancing in the paper.

The rest of the paper is organized as follows. In section \ref{sec: ConDesign} The conceptual design of the teaching manipulator is described. Section \ref{sec:ConfigD} proposes the configuration design. In section \ref{sec:StrucAnly}, balancing and operating force analysis and modeling are conducted, and condition of balancing and operating force is derived. In section \ref{sec:ProbFormu}, the objectives about and constraints are formulated. Section \ref{sec:MOO} explains the multi-objective optimization algorithm, and compares the optimal solutions with the original human expert design. An innovization study for the solutions is also conducted. Finally, we summarize the contributions and discuss some future work of the paper in section \ref{sec:conc}.

\section{Conceptual design}
\label{sec: ConDesign}
\begin{figure}
\centering
\includegraphics [width=6cm]{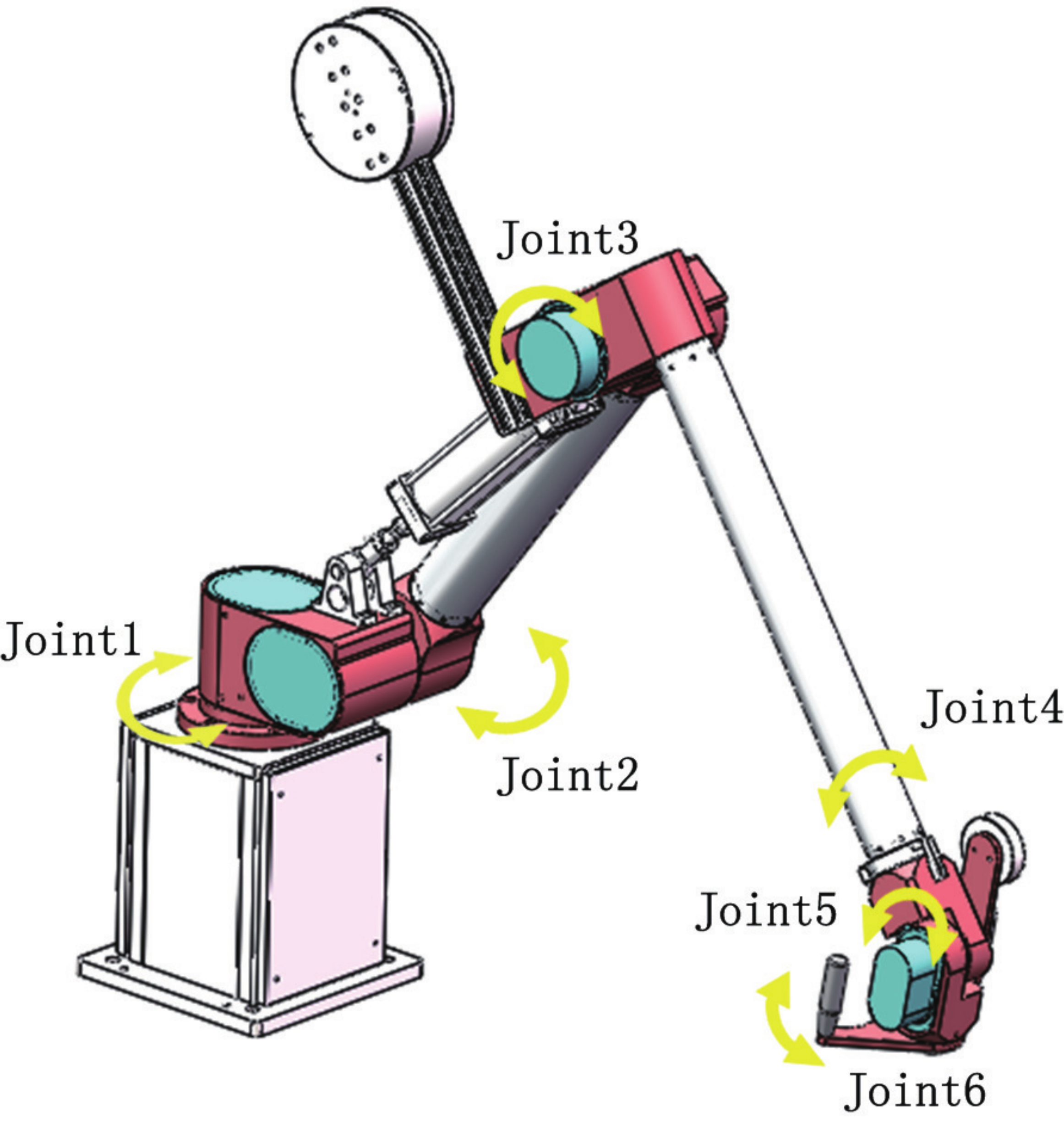}
\caption{A 6-DOF teaching manipulator}
 \label{JointPic}
\end{figure}

The teaching manipulator considered in this paper has six degrees of freedom (DOF), two DOF at the shoulder, one at the elbow and three at the wrist, as depicted in Fig. \ref{JointPic}. The link lengths of the manipulator are fixed. Six encoders are mounted for recording the angle position of each joint. It should be noted that there are no any actuators in the manipulator, which means that operators need to control it manually in the teaching procedure. To reduce the load of the operators, some structures are specially designed, including two counterweights at Joint 3 and 5, one pneumatic balancer at Joint 2 and three fiction disks inside Joint 1-3, in order to keep gravity balance in as many positions as possible in the workspace.

\begin{figure}
\centering
\includegraphics [width=8cm]{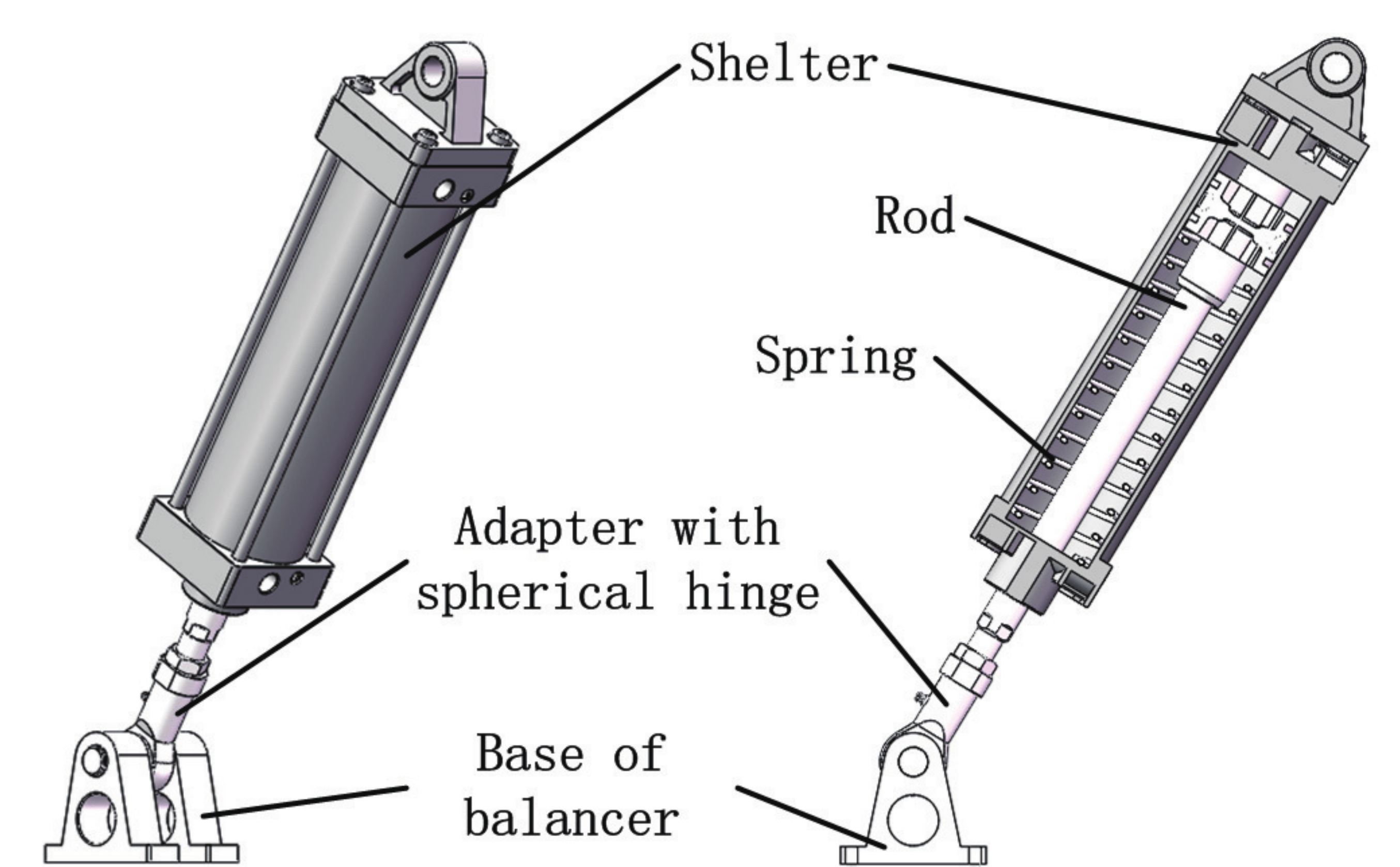}
\caption{The structure of balancer}
 \label{BalancerPic}
\end{figure}

To avoid requiring large space for rotation, instead of a counterweight at Joint 2, a pneumatic balancer is equipped. The balancer is an pneumatic cylinder with a spring mounted inside, as depicted in Fig. \ref{BalancerPic}. When the device works, the pneumatic cylinder needs to connect with an air pump with stable pressure. Then the balancer can provide a constant force to press the spring. The combination of the cylinder and the spring force is the drafting force against the loads. The drafting force can be considered as a linear force, proportional to the extension length of the balancer.

\begin{figure}
\flushleft
\includegraphics [width=8cm]{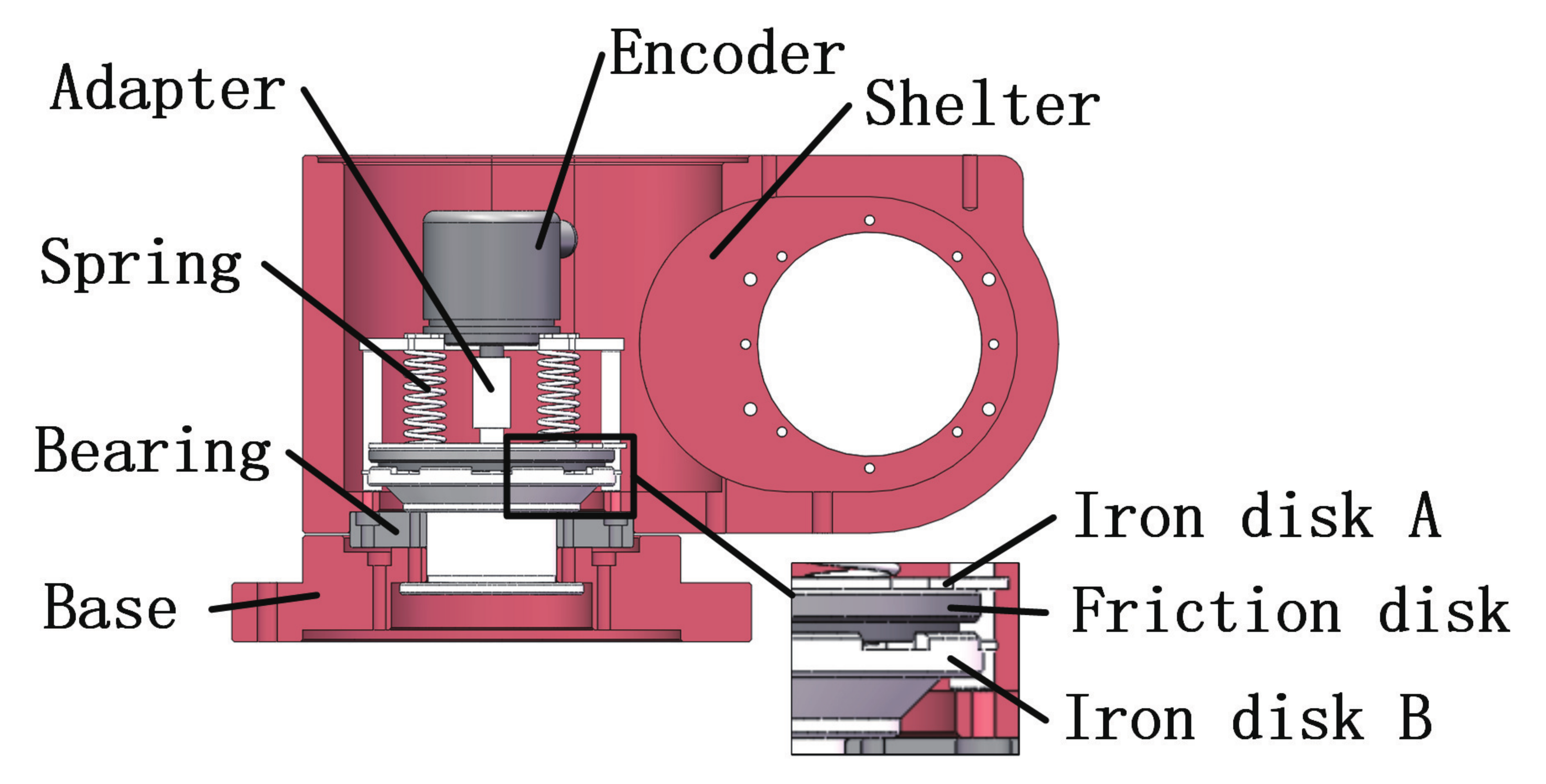}
\caption{The inner structure of Joint 1}
 \label{FicDPic}
\end{figure}

The counterweights and balancer alone may still fail to provide a satisfactory balancing effect in some positions. That is why the friction disks are set in Joint 1-3, providing extra resisting moments to address the imbalance problem adaptively. The physical realization of Joint 1 is illustrated in Fig. \ref{FicDPic}. The springs press the iron disks and the friction disk set in the middle. The friction between the disks provides the resisting moment against the imbalance. The friction moment can be adjusted via changing the pressing force of the springs.

\section{Configuration Design}
\label{sec:ConfigD}
The paper discusses optimizing the design of a kind of teaching manipulator using multi-objective evolutionary algorithm. According to the industrial requirement, the structure should ideally keep gravity balance in any position in its workspace, which is treated as a constraint. The multi-objective optimization considers the following three conflicting objectives to minimize: 1)\ \ the total mass of the device, 2)\ \ the maximal magnitude of the operating force and 3)\ \ the maximal difference between the maximal and the minimal operating force in a representative trajectory.

\subsection{Design variables}
\begin{figure}
\centering
\includegraphics [width=8.2cm]{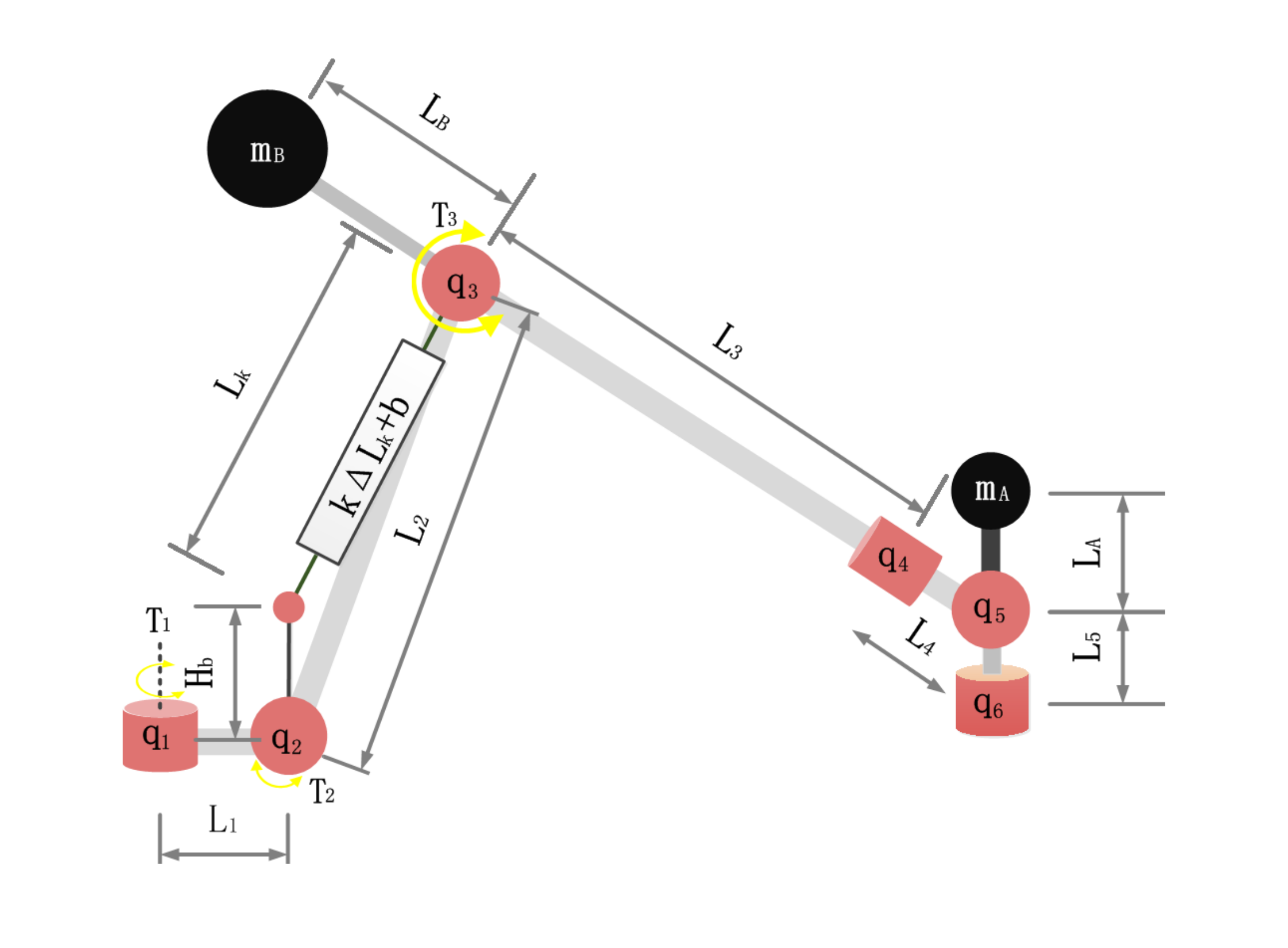}
\caption{Design variables and related parameters}
 \label{ParamPic}
\end{figure}

Nine design variables are considered in the optimization problem, which is
\begin{equation}
  X = (m_A, m_B, L_A, L_B, k, H_b, T_1, T_2, T_3)^T
\nonumber
\end{equation}
  where $m_A$, $m_B$ denote the masses of the two counterweights,  $L_A$, $L_B$ denote the distances between the centres of the mass of the counterweights and the rotation axis of the joints, namely, the lengths of the connecting rods of the counterweights. $k$ is the stiffness coefficient of the spring inside the balancer, $H_b$ is the length of a vertical virtual link between the lower attachment point and the rotation axis of Joint 2.  $T_i$ (i = 1, 2, 3 ) is the torque needed to overcome the moment of the disk friction and rotate Joint i. In gravity balancing condition, $T_i$ is approximately equal to the moment of the disk friction, therefore we treat $T_i$ as the moment of the disk friction. A schematic of the configuration of variables is shown in Fig. \ref{ParamPic}.

Other related parameters are also illustrated in Fig. \ref{ParamPic}. $m_i$ is the mass of Joint i, $m_{Li}$ and $L_i$ is the mass and the length of Link i. It is assumed that the density of each link is uniform, therefore the centres of mass are in the middle of the links. The mass per unit length of the connecting rod are $\rho_A$ and $\rho_B$, respectively.

\subsection{D-H parameters}
\begin{figure}
\centering
\includegraphics [width=8.2cm]{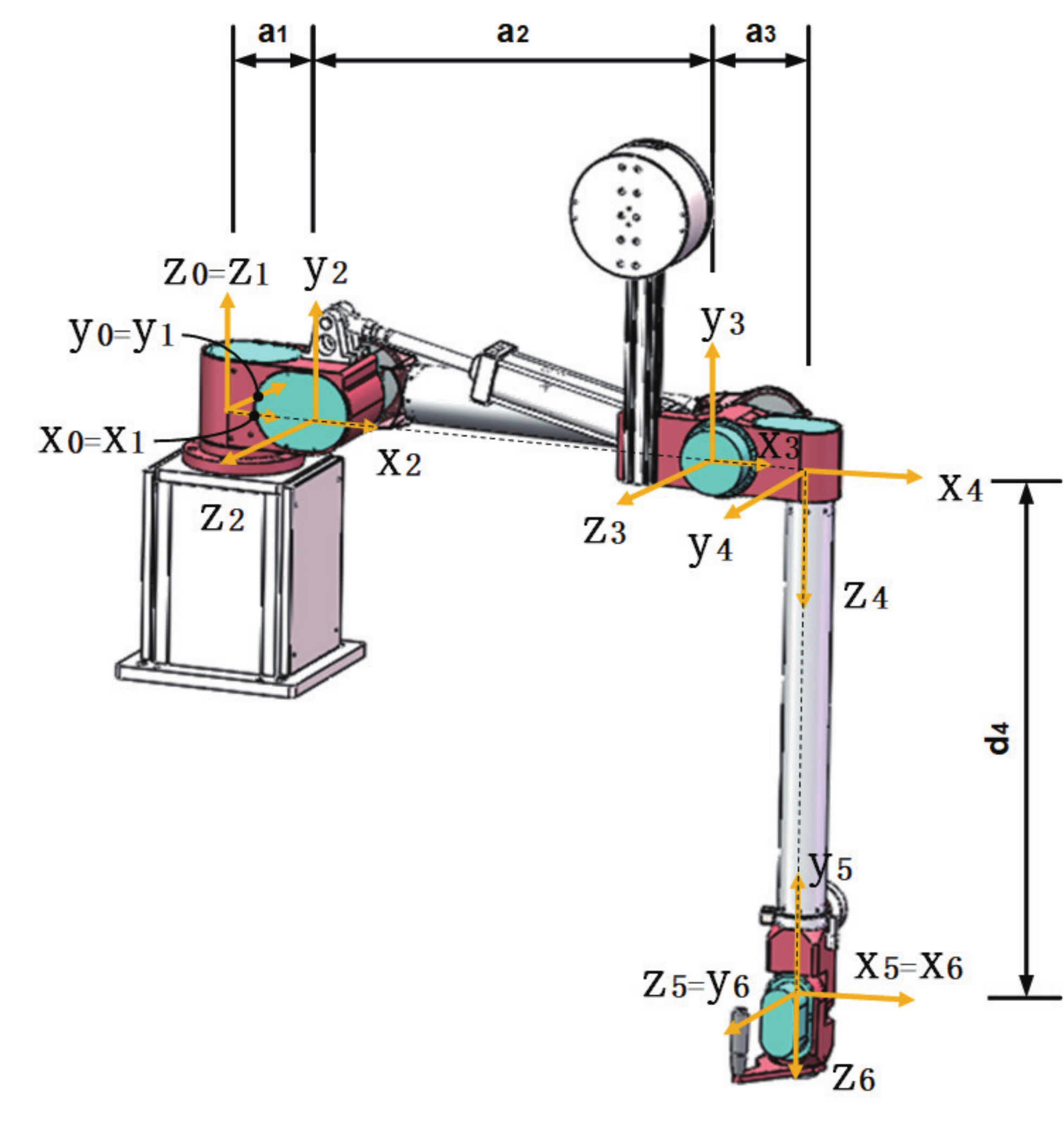}
\caption{Manipulator coordinate system}
 \label{DHPic}
\end{figure}

The forward kinematics of the teaching manipulator is formulated based on the Denavit-Hartenberg (D-H) convention \cite{denavit1955kinematic}. The coordinate frames $o_ix_iy_iz_i$ $(i = 1,2,\cdots,6)$ are assigned based on the sketch of the manipulator, which is shown in Fig. \ref{DHPic}. The D-H parameters are defined and listed in Table \ref{Table1}. We assume that the end effector and Joint 6 share the same coordinate frame.

 \begin{table}[ht]
 \small\sf\centering
 \caption{D-H Parameters of the robotic arm.\label{Table1}}
 \begin{tabular}{lllll}

 \toprule
 $Joint_i$&$\alpha_i$&$a_i$&$d_i$&$\theta_i$\\
 \midrule
 1 & $\pi/2$  &0.160 &0     &$q_1$\\
 2 & $0$      &0.790 &0     &$q_2$\\
 3 & $\pi/2$  &0.155 &0     &$q_3$\\
 4 & $-\pi/2$ &0     &0.995 &$q_4$\\
 5 & $\pi/2$  &0     &0     &$q_5$\\
 6 & $0$      &0     &0     &$q_6$\\
 \bottomrule
 \end{tabular}\\[10pt]
 \end{table}

\section{Structure analysis}
\label{sec:StrucAnly}

\subsection{ Balancing }
In the design, two counterweights, one balancer and three friction disks are used for balancing. Three joints with rotational axis, namely, Joint 2, 3 and 5, which are parallel to the horizontal plane are needed to design the balancing structure. There is an extra friction disk setting in Joint 1 for the adjustment of operating force performance.
Let $G_j$ be the gravitational moment reacting at the axis of Joint j, while $P_j$ be the moment provided by the counterweight and its connecting rod or the balancer, reacting at the axis of Joint j.

At Joint 5, a counterweight is designed for balancing the gravity of the Link 5 and Joint 6. The masses of the counterweight and the connecting rod, with homogeneous density are considered. Here we have the balancing equation, which is $G_5 = P_5$. However, it is difficult to satisfy an equality constraint. In industrial reality, it is not necessary to making it equal, because the operator can handle small imbalance in the wrist through the manually operable handgrip. Therefore, under $5\%$ imbalance is allowed. Then we can specify this using an inequality constraint, which is
\begin{equation}
| G_5 - P_5 | \leq 5\% G_5
\label{balance5}
\end{equation}
where
\begin{equation}
G_5 = ( m_6 + m_{L5})gL_5\cos{q_5}
\label{G5}
\end{equation}
\begin{equation}
P_5 = (m_A gL_A + \frac{L_A^2 \rho_A g}{2})\cos{q_5}
\label{P5}
\end{equation}
where $m_6$, $m_{L5}$ are the masses of Joint 6 and Link 5, $L_5$ is the length of Link 5, $\rho_A$ is linear density of the connecting rod, which connects counterweight A and the shelter of Joint 5, $g$ is the gravitational acceleration and we take $g = 9.8 m/s$.

At Joint 3, a counterweight and a friction disk are used for balancing the gravity. The friction disk can provide a static frictional moment which correspondingly increases as the imbalance moment increased, until the disk starts to rotate. If the disk rotates, the moment becomes a kinetic frictional moment, which can be treated as a constant. Because the disk moves with a low velocity, the kinetic frictional moment is approximately equal to the static frictional moment. Therefore, we have the following balance equation, which is
\begin{equation}
\max_{q_3}{| G_3 - P_3 | \leq T_3}
\label{balance3}
\end{equation}
where
\begin{equation}
\begin{split}
G_3 =& \{(m_A + m_6 + m_{L5} + m_5 + L_A \rho_A)(L_3 + L_4)g \\
 &+ m_4 g L_3 + \frac{m_{L3} g L_3}{2}+ \frac{(2L_3+L_4) m_{L4} g}{2}\}\cos{q_3}
\end{split}
\end{equation}
\begin{equation}
P_3 = (m_B g L_B + \frac{L_B^2 \rho_B g}{2})\cos{q_3}
\end{equation}
where $m_{L3}$ is the mass of Link 3.

\begin{figure*}
\centering
\includegraphics [width=12cm]{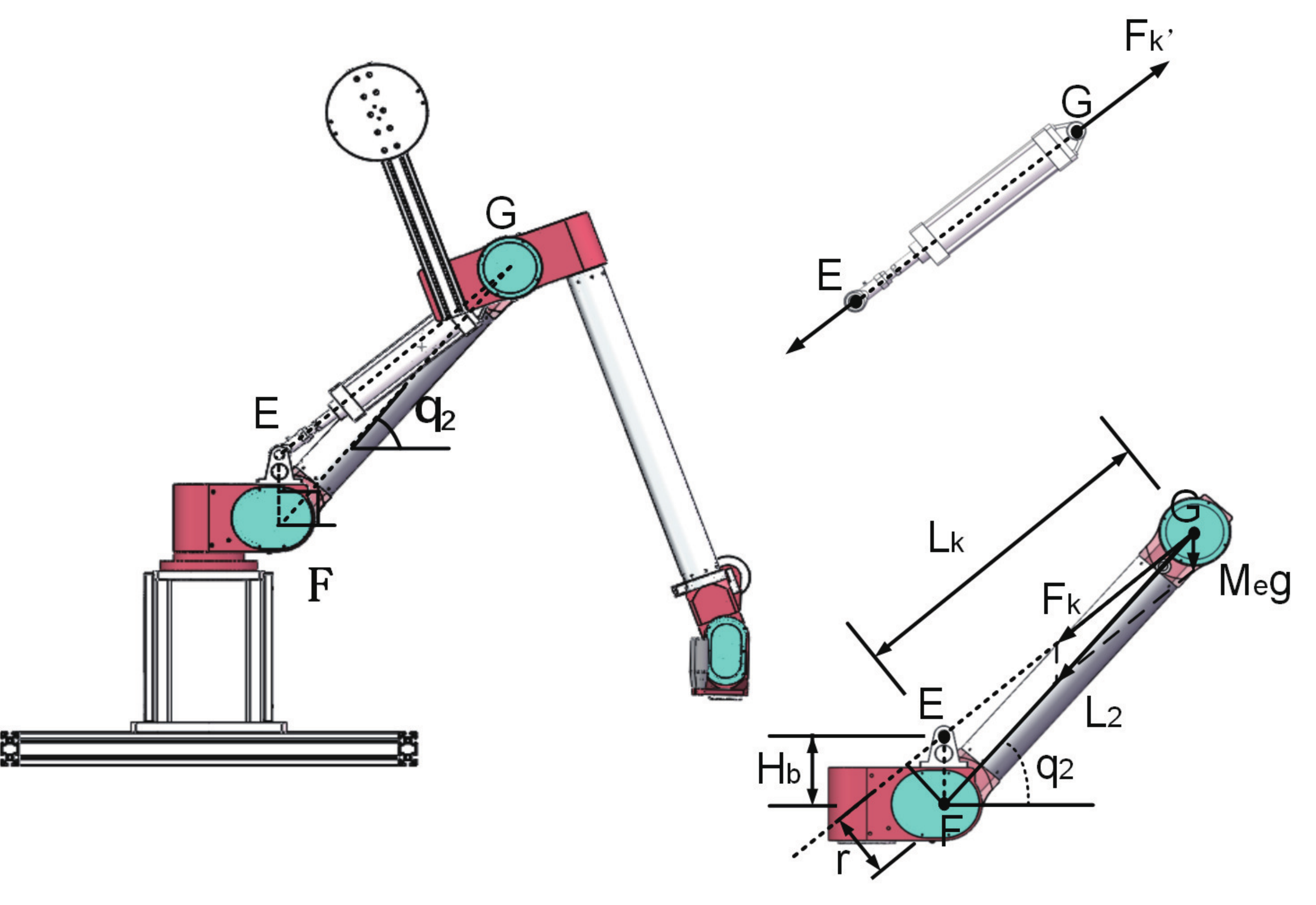}
\caption{Force analysis of Link 2}
\label{ForceAnalysis}
\end{figure*}

At Joint 2, a balancer and a friction disk are mounted for balancing. As Link 2 rotates about Joint 2, the balancer, Link 2 and a virtual link EF make up a triangle $\bigtriangleup EFG$, which is shown in Fig. \ref{ForceAnalysis}. $L_k$ is the total length of the balancer. $F_k$ is the drafting force of the balancer in a given position, which can balance the equivalent mass $M_n$ acting on point G. We can figure out the length of the balancer based on the triangle relation, which is
\begin{equation}
L_k=\sqrt {H_b^2+L_2^2-2H_bL_2\cos{(\pi/2-q_2)}}
\label{lk}
\end{equation}
where $q_2$ is the rotation angle of Joint 2. The drafting force of the balancer is linearly proportional to the variation of the balancer length. The shortest length of the balancer appears when Link 2 is in vertical position. In this situation, $\bigtriangleup EFG$ becomes a line and the length of the balancer is
\begin{equation}
L_{k0} = L_2 - H_b
\label{lk0}
\end{equation}
Thus,
\begin{equation}
F_k=k(L_k-L_{k0})-b
\label{fk}
\end{equation}
where b is the constant force provided by the cylinder.

In Fig. \ref{ForceAnalysis}, there are three forces reacting at Link 2. $F_E$ points along Link 2. Only when the resultant force of $F_k$ and the equivalent gravity $M_eg$ are aligned with Link 2, Link 2 can be in equilibrium. From the law of sine, we have the relation that
\begin{equation}
\frac{\sin{(\pi/2-q_2)}}{L_k} = \frac{\sin{\alpha}}{H_b}
\label{Lawofsine}
\end{equation}
where $\alpha$ is the angle of $\angle FGE$, shown in Fig. \ref{ForceAnalysis}.
Thus, the moment provided by the balancer is
\begin{equation}
P_2=L_2F_k\sin{\alpha}
\label{P2base}
\end{equation}
With Eq.(\ref{lk})-(\ref{Lawofsine}), we can simplify Eq.(\ref{P2base})
\begin{equation}
\begin{split}
P_2=k H_b L_2 \cos{q_2}
+\frac{(k(H_b-L_{2})-b ) H_b L_2 \cos{q_2}}{\sqrt {H_b^2+L_2^2-2H_bL_2\sin{q_2}}}
\label{P2sim}
\end{split}
\end{equation}

Because the mass of Link 2 is homogeneous, the gravitational moment of Link 2 is $\frac{1}{2}m_{L2}gL_2$, which is equivalent to the moment produced by half of $m_2$ at point G. The mass of the balancer is not considered in the structure because it is much smaller than the equivalent payload acting at point G. Thus, we can formulate $G_2$, which are
\begin{equation}
\begin{split}
&G_2 = M_e g L_2 \cos{q_2}
\end{split}
\end{equation}
$M_e$ is the equivalent mass.
\begin{equation}
\begin{split}
M_e = &\sum_{i=3}^6 m_i + m_A + m_B + L_A \rho_A + L_B \rho_B \\
&+ m_{L3} + m_{L4}+ m_{L5} + \frac{1}{2}m_{L2}\
\label{me}
\end{split}
\end{equation}
Therefore, the value of the imbalanced moment can be formulated as follows.
\begin{equation}
\begin{split}
|G_2-P_2|=&|(M_eg -kH_b)L_2 \cos{q_2} \\
&+\frac{(k(H_b-L_{2})-b) H_b L_2 \cos{q_2}}{\sqrt {H_b^2+L_2^2-2h_bL_2\cos{q_2}}}|
\label{balancerForce}
\end{split}
\end{equation}

We expect to minimize the operating force with balancing constraints, so the balancing condition without fiction disk should be considered first because the operating force is zero in an ideal situation. Then we introduce the disk friction to keep balance in non-ideal cases.
The ideal gravity balance conditions of teaching manipulator should be independent to the position of the joints. Thus, weather the equality
\begin{equation}
|G_2-P_2|=0
\label{J2}
\end{equation}
is satisfied should be independent to $q_2$.

Observe Eq.(\ref{balancerForce}), only when
\begin{equation}
k(L_2 - H_b)-b=0
\label{condition1}
\end{equation}
\begin{equation}
  M_eg - kH_b=0
\label{condition2}
\end{equation}
$cos{q_2}$ is eliminated and Eq.(\ref{J2}) is satisfied in any rotational angle.

Now we consider the real situation with error. Because of the errors of manufacturing and assembly, the device usually fails to keep balance, therefore the fiction disk is mounted for keep gravity balance from the error. Thus, we have
\begin{equation}
|G_2 - P_2| \leq T_2
\label{balance2}
\end{equation}
With the errors, Eq.(\ref{condition1}) can be still satisfied by adjusting the force of the pneumatic cylinder $b$, however, Eq. (\ref{condition2}) is difficult to be satisfied because the parameters are all related to manufacturing and assembly.  Thus, we have
\begin{equation}
\begin{split}
|G_2 - P_2| =&|(M_eg -kH_b)L_2 \cos{q_2} |\\
=&|M_eg -kH_b|L_2 |\cos{q_2}|
\end{split}
\end{equation}
The maximal value of $|G_2 - P_2|$ can be obtained when $|\cos{q_2}| = 1$. Thus, we have
\begin{equation}
\begin{split}
\max_{q_2}{|G_2 - P_2|} =|M_eg - kH_b|L_2 \leq T_2
\label{balance2sim}
\end{split}
\end{equation}

\subsection{Total mass of the device}
Minimizing the total mass of the device is expected, because lighter weight usually means lower cost and lower power consumption. The mass of the six joints, five links, two counterweights and their connecting rods are all considered. Here we have the total mass equation as follows.
\begin{equation}
M = \sum_{i=1}^6 m_i + \sum_{p=1}^5 m_{Lp} + m_A + m_B+ L_A \rho_A + L_B \rho_B
\label{totomass}
\end{equation}

\subsection{Operating force analysis}
It is expected that the manipulator can be manually controlled using minimal operating force, and with minimal variation of the operating force along a trajectory. We need to analyze the relation between the operating force and the joint moments.

Let $[F] = [f_x, f_y, f_z, m_x, m_y, m_z]^T$ be the spatial force vector of the end effector in the end-effector frame, where $f_r$ $(r = x,y,z)$ is the force along r axis and $m_r$ is the moment about r axis. Let $[\tau] = [\tau_1, \tau_2, \tau_3, \tau_4, \tau_5, \tau_6]^T$ be the torque vector of the six joints, where $\tau_i$ $(i = 1,2,3,\cdots,6)$ is the torque at Joint i. Here we have the equation,
\begin{equation}
[\tau] = [J]^T[F]
\end{equation}
where $[J]^T$ is the Jacobian matrix in the base coordinate system. To generate a moment and overcome the friction moment $[\tau] = [T_1, T_2, T_3,0,0,0]^T$, an operating force and moment $[F]$ is needed, which is
\begin{equation}
[F] =([J]^T)^{-1}[\tau]
\label{ctrlforceV}
\end{equation}
When the end effector moves along a trajectory, we equally divide the trajectory into N segments. We can figure out $[J]$ for each segment, and then the the spatial force vector $[F]$. Because the operating moment reacts at the wrist with small load, only the module of operating force $F_c(X,t)$ is considered in this case. Thus,
\begin{equation}
F_c(X,t) = \sqrt {f_x^2(x,t) + f_y^2(x,t) + f_z^2(x,t)}
\label{ctrlforce}
\end{equation}

\section{Problem formulation}
\label{sec:ProbFormu}
The goal of the optimization problem is to optimize objective functions simultaneously by satisfying the gravity balance constraints. The vector of nine design variables is $X = (m_A, m_B, L_A, L_B, k, H_b, T_1, T_2, T_3)^T$, where  $m_A$, $m_B$, $L_A$, $L_B$ are the variables about the two counterweights, $k$, $H_b$ are about the balancer and the rest are about the three friction disks. Details of objective functions and constraints will be given in the following subsections.
\subsection{Objective functions}
Three objectives are considered, which are to minimize the total mass of the  device, the maximal operating force and the maximal difference of the maximal and minimal operating force in a trajectory.
\subsubsection{Total mass}
Based on Eq.(\ref{totomass}), the objective function can be written as
 \begin{equation}
f_1(x)=M
\end{equation}

\subsubsection{Maximal operating force}
It is a bilevel optimization problem, which contains two levels of optimization tasks \cite{sinha2017evolutionary}. Bilevel optimization problem is difficult to handle, because only the optimal solutions of the lower level optimization problem are considered as feasible candidates of the upper level optimization problem.
We prefer to minimize the maximal operating force in a trajectory to ensure that an operator can easily maneuver the device. Based on Eq.(\ref{ctrlforceV}) and Eq.(\ref{ctrlforce}), the objective function can be written as
 \begin{equation}
f_2(x)=\max_tF_c(X,t)
\end{equation}

\subsubsection{Variation of operating force}
Large variation of operating force can disrupt a normal operation. The third objective is to minimize the difference between the maximum and minimum of operating force, which can be formulated as
 \begin{equation}
f_3(x)=|\max_tF_c(X,t)-\min_tF_c(X,t)|
\end{equation}

Again, the third objective indicateds a bi-level optimization problem. Because of the bilevel optimization formulas, the optimal solutions are not easy to be discovered in traditional step-by-step design procedure. As a result, muiti-objective optimization algorithm is used in this work.

\subsection{Constraints}
The constraints are mainly about balance conditions and the bounds of the design variables. The constraints can be derived as follows:
 \begin{equation}
| G_5 - P_5 | \leq 5\% G_5
\nonumber
\end{equation}
\begin{equation}
\max_{q_3}{| G_3 - P_3 | }\leq  T_3
\nonumber
\end{equation}
\begin{equation}
\max_{q_2}{|G_2 - P_2|} \leq T_2
\nonumber
\end{equation}
which are Eq.(\ref{balance5}), Eq.(\ref{balance3}) and Eq.(\ref{balance2sim}), respectively.

The bounds of the design variables are listed in Table \ref{Table2}
\begin{table}[ht]
\small\sf\centering
\caption{The bounds of design variables.\label{Table2}}
\begin{tabular}{llll}
\toprule
Variables&Range&Units&Current value\\
\midrule
$m_A$ & [0.3, 20]  &kg & 1.6\\
$m_B$ & [19, 50]   &kg & 25\\
$L_A$ & [0.11, 0.5]&m  & 0.185\\
$L_B$ & [0.2 0.8]  &m  & 0.462\\
$k  $ & [0, 15000] &$N/m$ &3730\\
$H_b$ & [0.11, 0.18]&m   &0.15\\
$T_1$ & [0, 90]    &$N\centerdot m$ &75.7\\
$T_2$ & [0, 90]    &$N\centerdot m$ &75.7\\
$T_3$ & [0, 90]    &$N\centerdot m$ &75.7\\
\bottomrule
\end{tabular}\\[10pt]
\end{table}

Other constant parameters needed are listed in Table \ref{Table3}
\begin{table}[ht]
\small\sf\centering
\caption{The value of relative parameters.\label{Table3}}
\begin{tabular}{lll}
\toprule
Parameter&value&Units\\
\midrule
$m_1$ &4.136 &kg\\
$m_2$ &8.225 &kg\\
$m_3$ &9.665 &kg\\
$m_4$ &1.249 &kg\\
$m_5$ &4.185 &kg\\
$m_6$ &2.013 &kg\\
$m_{L1}$ &0.631 &kg\\
$m_{L2}$ &2.071 &kg\\
$m_{L3}$ &1.816 &kg\\
$m_{L4}$ &0.340 &kg\\
$m_{L5}$ &0.358 &kg\\
$L_2$  &0.79  &m\\
$\rho_A$ &1.8 &kg/m\\
$\rho_B$ &3.7 &kg/m\\
\bottomrule
\end{tabular}\\[10pt]
\end{table}

\subsection{Trajectory design}
To evaluate the objectives, the optimization problem needs a specific a trajectory. We define a representative trajectory of the end-effector in the base coordinate system, which is
\begin{equation}
\begin{split}
&X_{ef} =1.5 - 0.25(1 - \cos(t))\\
&Y_{ef}= - 0.25+0.5\times(1 - \cos(t/2))\\
&Z_{ef}=    0.5+0.5\times(\cos(t/2) - 1)
\nonumber
\end{split}
\end{equation}
all with unit of m. The Euler angle for the end-effector are given as $[0, \pi, \pi]$, which implies that the end-effector remains vertical and points at the ground during the prescribed motion. To figure out the Jacobian matrix $[J]$, we equally divide the trajectory into $N=500$ segments.

\subsection{Multi-objective design optimization}
\label{sec:MOO}
After formulating the optimization problem, the next step is to use an appropriate algorithm to search for an optimal design set. The optimization model is a multi-objective problem, and all of the constraints must be taken into account. Due to the complexity and size of the problem, NSGA-II-CDP is chosen to solve the optimization problem. This algorithm produces a Pareto front that consists of a set of optimal solutions, which are not dominated by each other and other solutions.

\subsection{Multi-objective optimization algorithm}
\begin{figure}
\centering
\includegraphics [width=8.5cm]{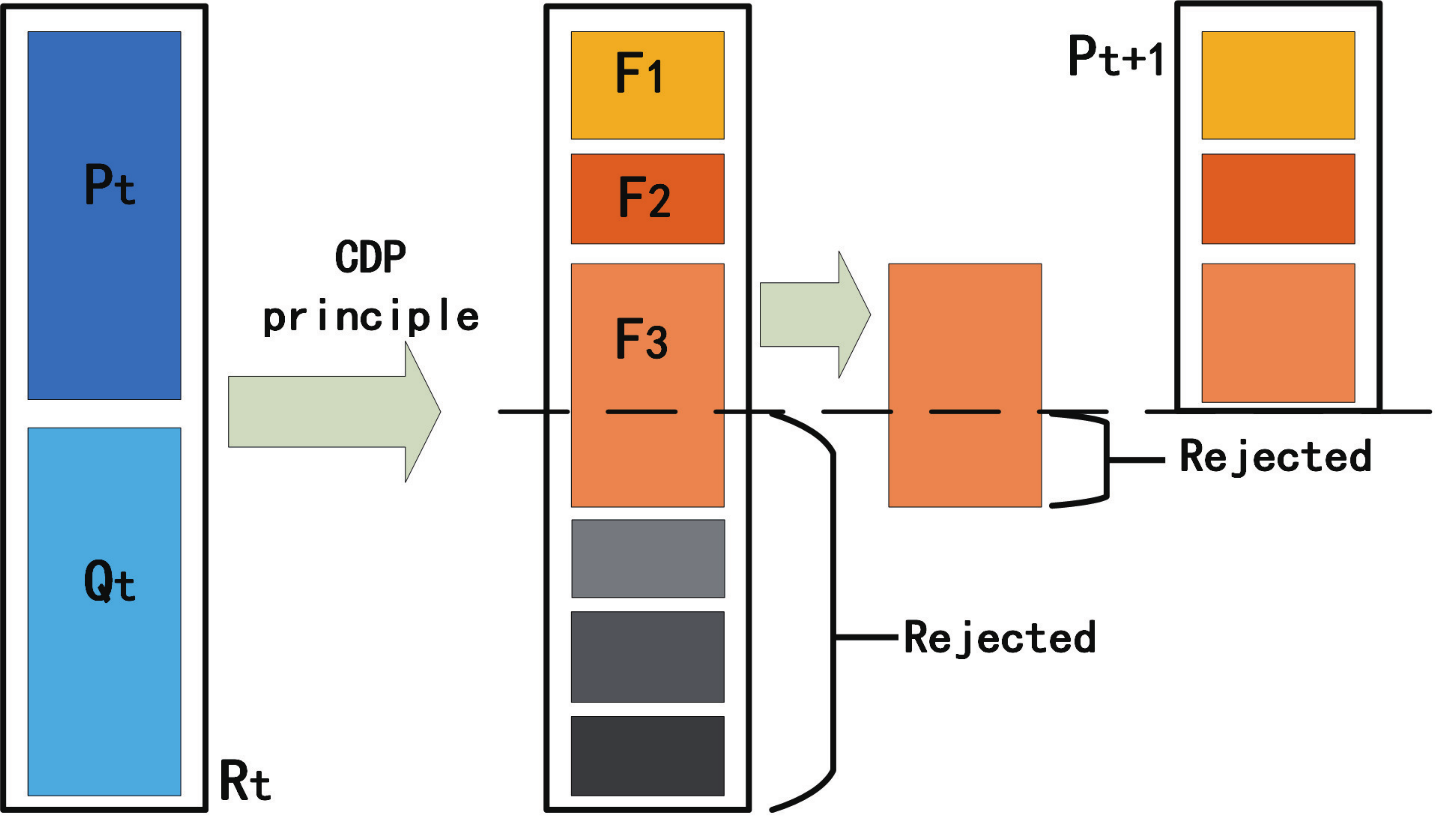}
\caption{NSGA-II-CDP procedure, modified from NSGA-II \cite{deb2002fast}}
 \label{NSGA}
\end{figure}

Fig. \ref{NSGA} is the illustration of the NSGA-II-CDP procedure. In this procedure, an offspring $Q_{t}$ is produced by the operator of crossover, mutation and selection from the working population $P_t$. Then, a new population $R_t$ is constructed by combing $P_t$ and $Q_t$. $R_t$ is sorted based on the CDP principle to divide the population into different fronts. The CDP principle is defined as follows:
\begin{enumerate}
  \item[(a)] In the case of that solution $i$ and $j$ are both feasible, the one dominating the other is better.
  \item[(b)] In the case that solution $i$ is  feasible and solution $j$ is infeasible, solution $i$ is better than solution $j$.
  \item[(c)] In the case that solution $i$ and $j$ are both infeasible, the one with smaller constraint violation is better than the other.
\end{enumerate}

Each solution is assigned to a non-dominated rank based on the CDP principle. Moreover, the crowding distance is calculated to sort the solutions in $f_3$. Then, the first $N$ solutions are selected to construct the population $P_{t+1}$.

 The parameters of NSGA-II-CDP are listed as follow:
\begin{enumerate}
  \item population size $=300$
  \item population size $=300$
  \item number of generations $=5000$
  \item crossover probability $=0.9$
  \item mutation probability $=1/n$, where $n=9$ is the dimension of variables
  \item the distribution parameter in the polynomial mutation is 20
  \item the distribution parameter in the simulated binary operator is 20
\end{enumerate}

\subsection{Result analysis}
\begin{figure}
\centering
\includegraphics [width=8cm]{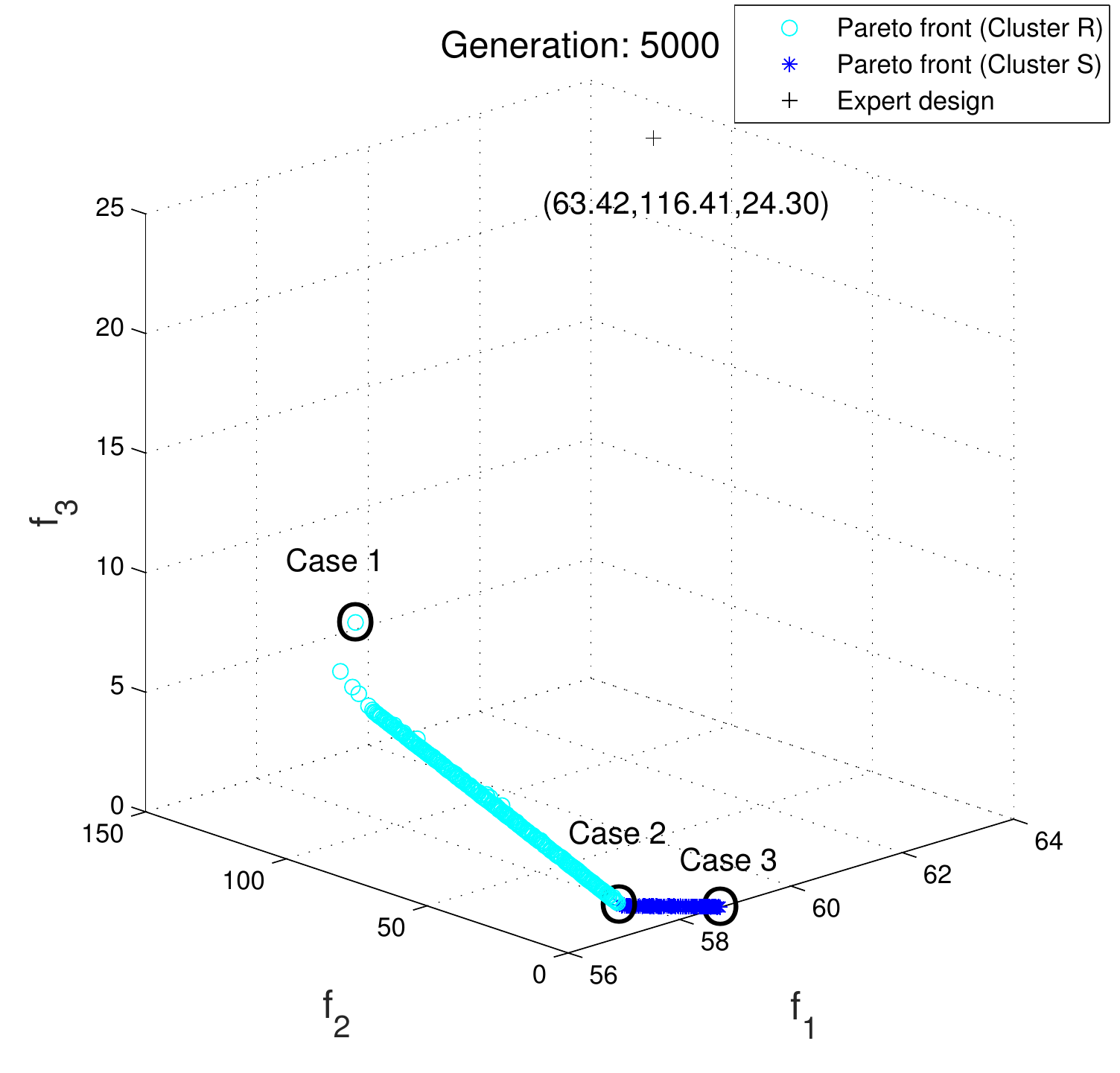}
\caption{The Pareto-front and the original design of a human expert is shown in the objective space, with three cases circled out. The obtained solutions are all better than the expert design in terms of the three objectives.}
 \label{VSorg}
\end{figure}

The non-dominated solutions, namely Pareto front, is shown in Fig. \ref{VSorg}. It is easy to notice that $f_1$ $f_2$ and $f_3$ are conflicting. The Pareto-front is combined by two clusters, which are cluster R with less than 57.33 total mass and cluster S with no less than 57.33 total mass. Each of them is nearly in a straight line, with Eq.(\ref{result1}) and (\ref{result2})

\begin{equation}
f_2=\left\{\begin{array}{ll}
-59.84f_1+3438.5& f_1\in[56.22,57.33], \\
-5.40f_1+317.27 & f_1\in[57.33,58.72],
\end{array}\right.
\label{result1}
\end{equation}
\begin{equation}
f_3=\left\{\begin{array}{ll}
-5.80f_1+333.43&f_1\in[56.22, 57.33],\\
-0.52f_1+30.631&f_1\in[57.33, 58.72],
\end{array}\right.
\label{result2}
\end{equation}

\begin{figure*}
\begin{tabular}{cccc}
\begin{minipage}[t]{0.23\linewidth}
\includegraphics[width = 3.6cm]{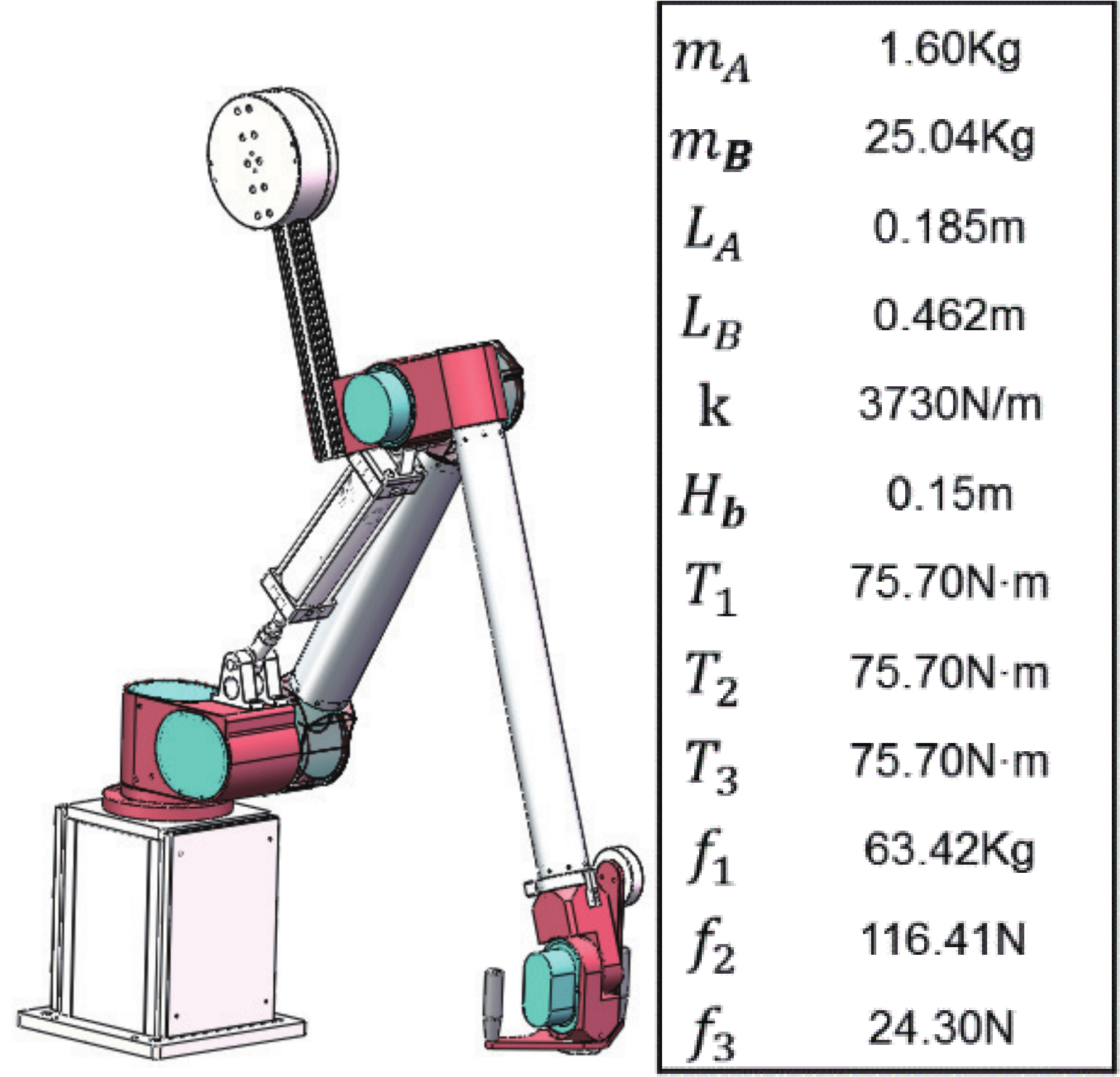} \\
\centering{\scriptsize{(a) Expert design}}
\end{minipage}
\begin{minipage}[t]{0.23\linewidth}
\includegraphics[width = 3.6cm]{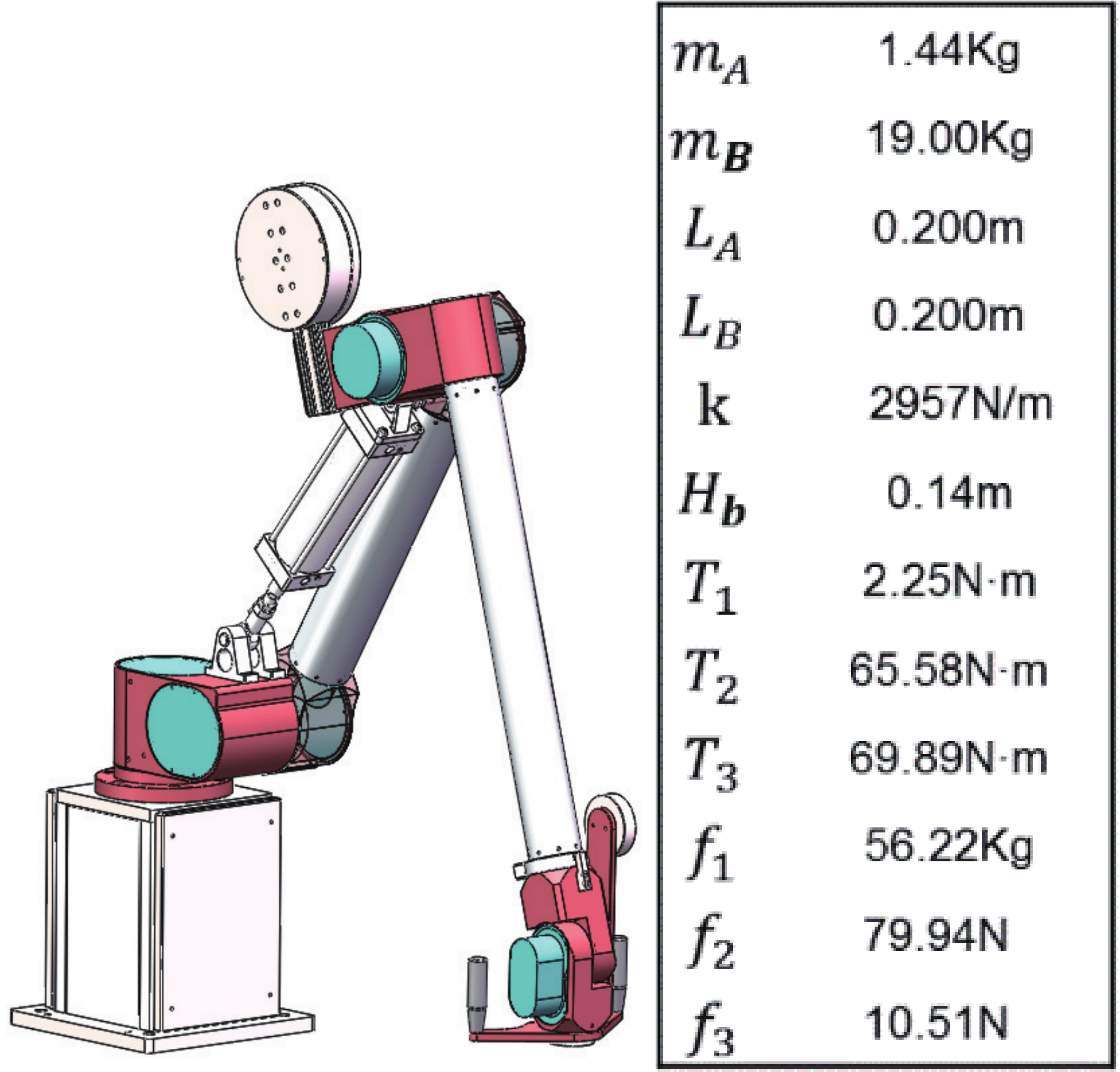}\\
\centering{\scriptsize{(b) Optimal solution 1}}
\end{minipage}
\begin{minipage}[t]{0.23\linewidth}
\includegraphics[width =3.6cm]{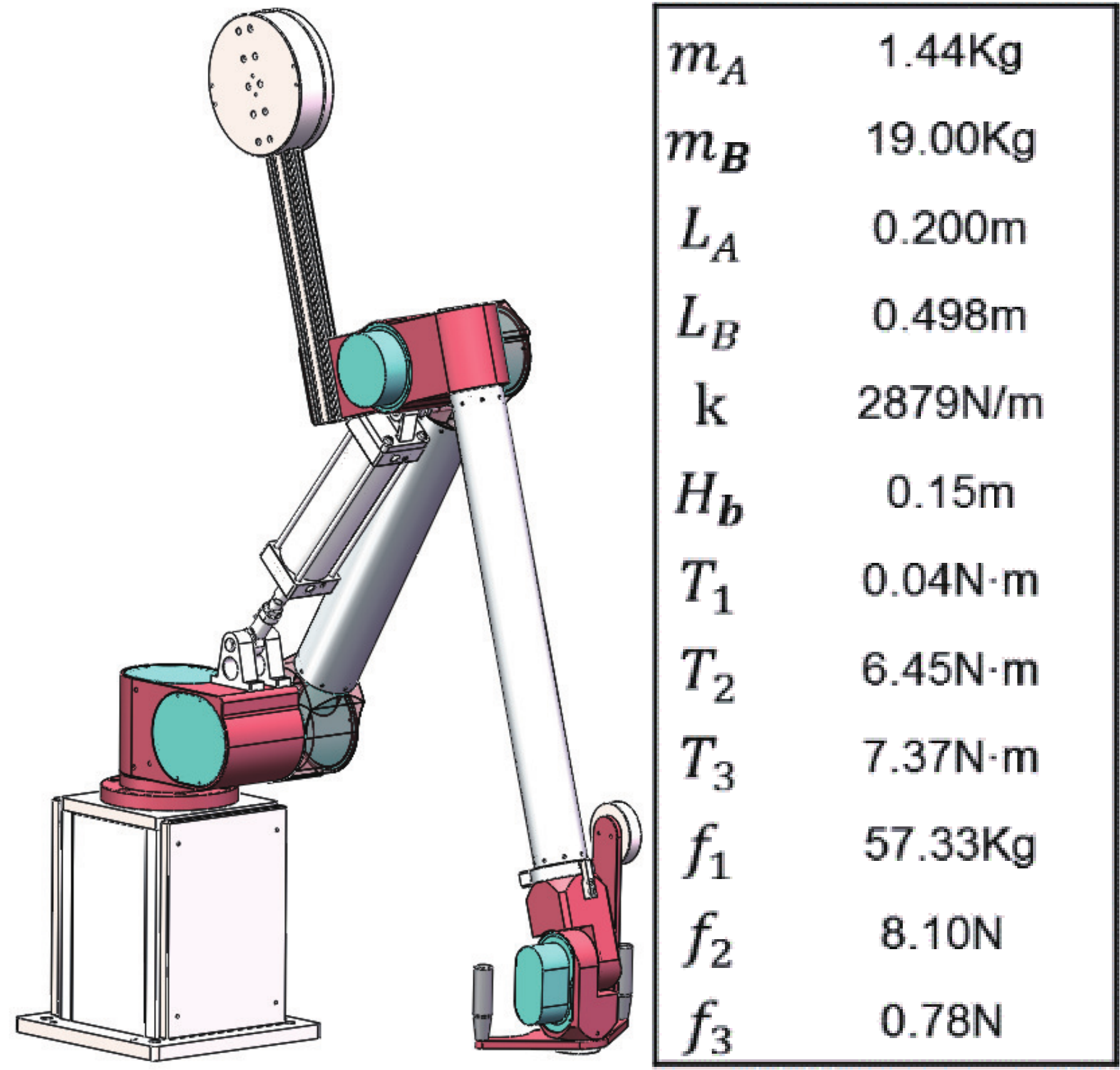}\\
\centering{\scriptsize{(c) Optimal solution 2}}
\end{minipage}
\begin{minipage}[t]{0.23\linewidth}
\includegraphics[width =3.6cm]{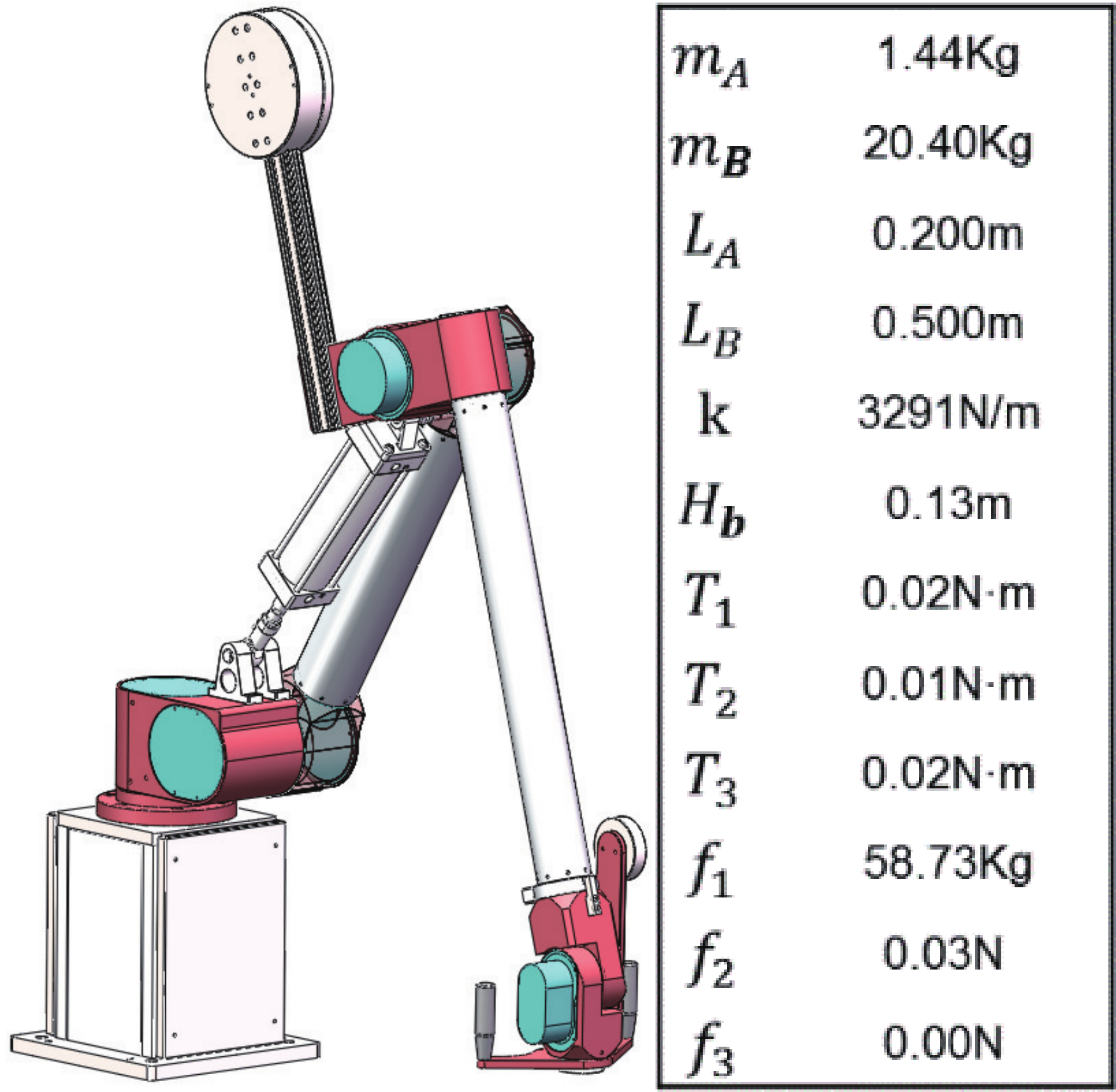}\\
\centering{\scriptsize{(d) Optimal solution 3}}
\end{minipage}
\end{tabular}
\caption{3D model of original design and obtained solution}
\label{Model}
\end{figure*}

The obtained solutions are all better than the expert design in terms of the three objectives. Two extreme points and one turning point are chosen from the Pareto front as samples, as shown in Fig. \ref{VSorg}. Their designs are compared with the original design of a human expert,  with the design variables and the three objectives listed in Fig. \ref{Model}.

Based on preferences of the three objectives, the users can choose a proper solution on the Pareto-optimal front. Meanwhile, it is worthy to notice that the appearance of three solutions are different, while the friction moments of disks inside the three joints are also different. The differences and the interaction among the valuables make the design better. There might be some principles or meaningful relationships behind the solution data. Therefore, an innovization study is conducted.

\subsection{Innovization study}
\begin{figure*}
\begin{tabular}{cc}
\begin{minipage}[t]{0.45\linewidth}
\centering
\includegraphics [width=7.5cm]{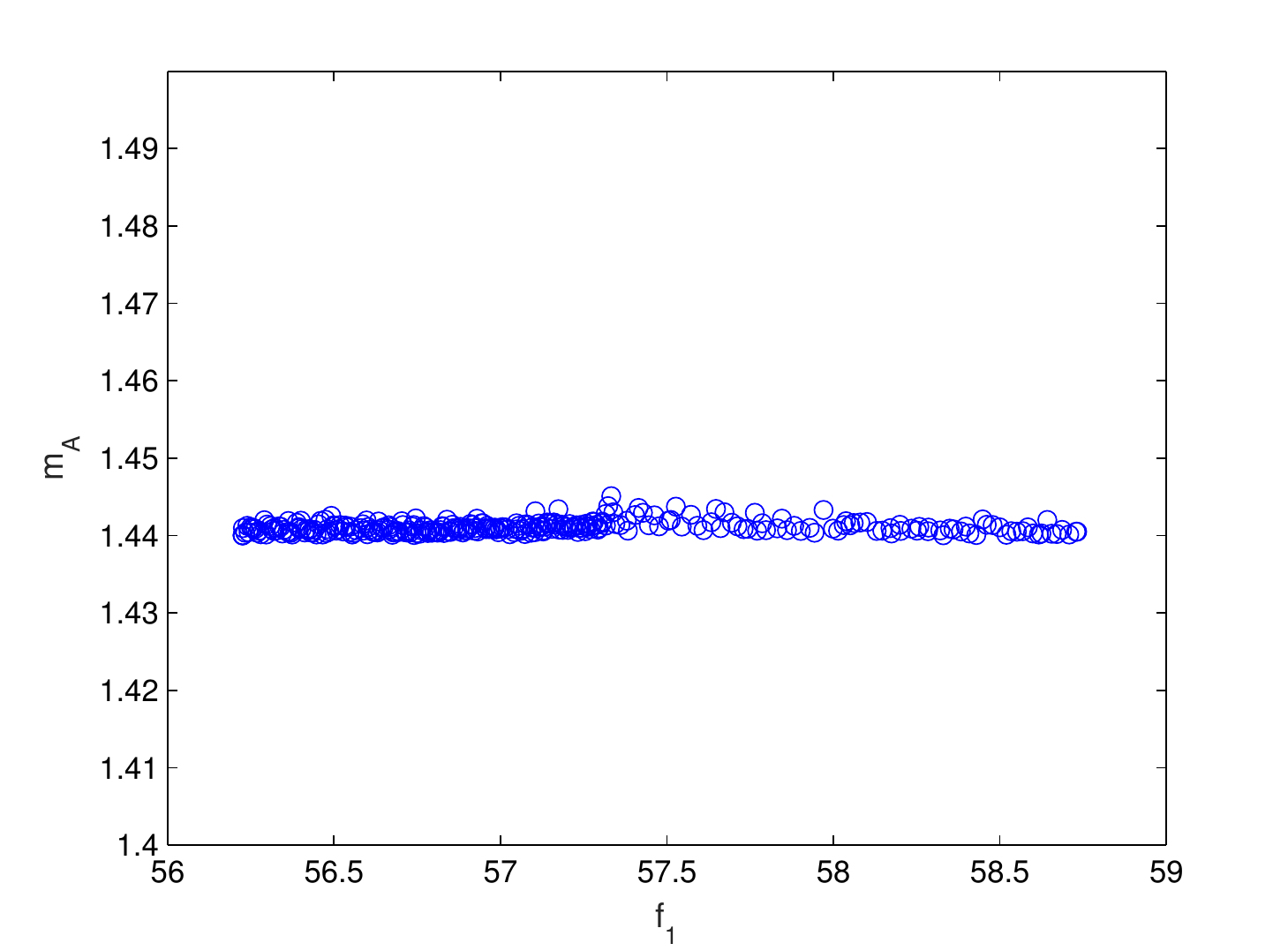}
\caption{Variation of the mass of Counterweight A $m_A$ with the total mass is shown. $m_A$ is mostly fixed at about 1.44 kg.}
 \label{mArep}
\end{minipage}
\hspace{1.2cm}
\begin{minipage}[t]{0.45\linewidth}
\centering
\includegraphics [width=7.5cm]{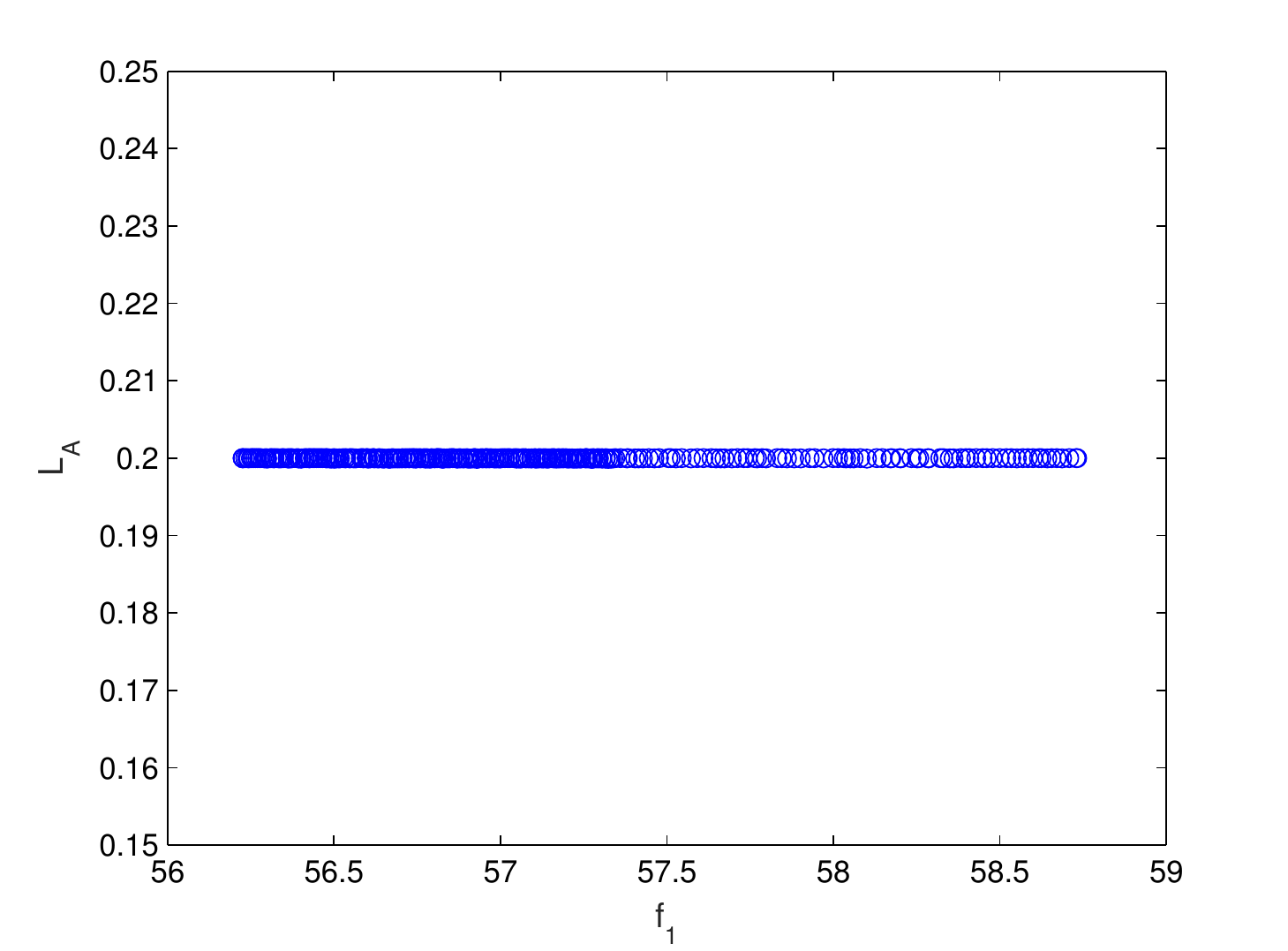}
\caption{Variation of the length of Counterweight A connecting rod $L_A$ with the total mass is shown. $L_A$ is fixed at about 0.2, which is the upper bound.}
 \label{LArep}
\end{minipage}
\end{tabular}

\begin{tabular}{cc}
\begin{minipage}[t]{0.45\linewidth}
\centering
\includegraphics [width=7.5cm]{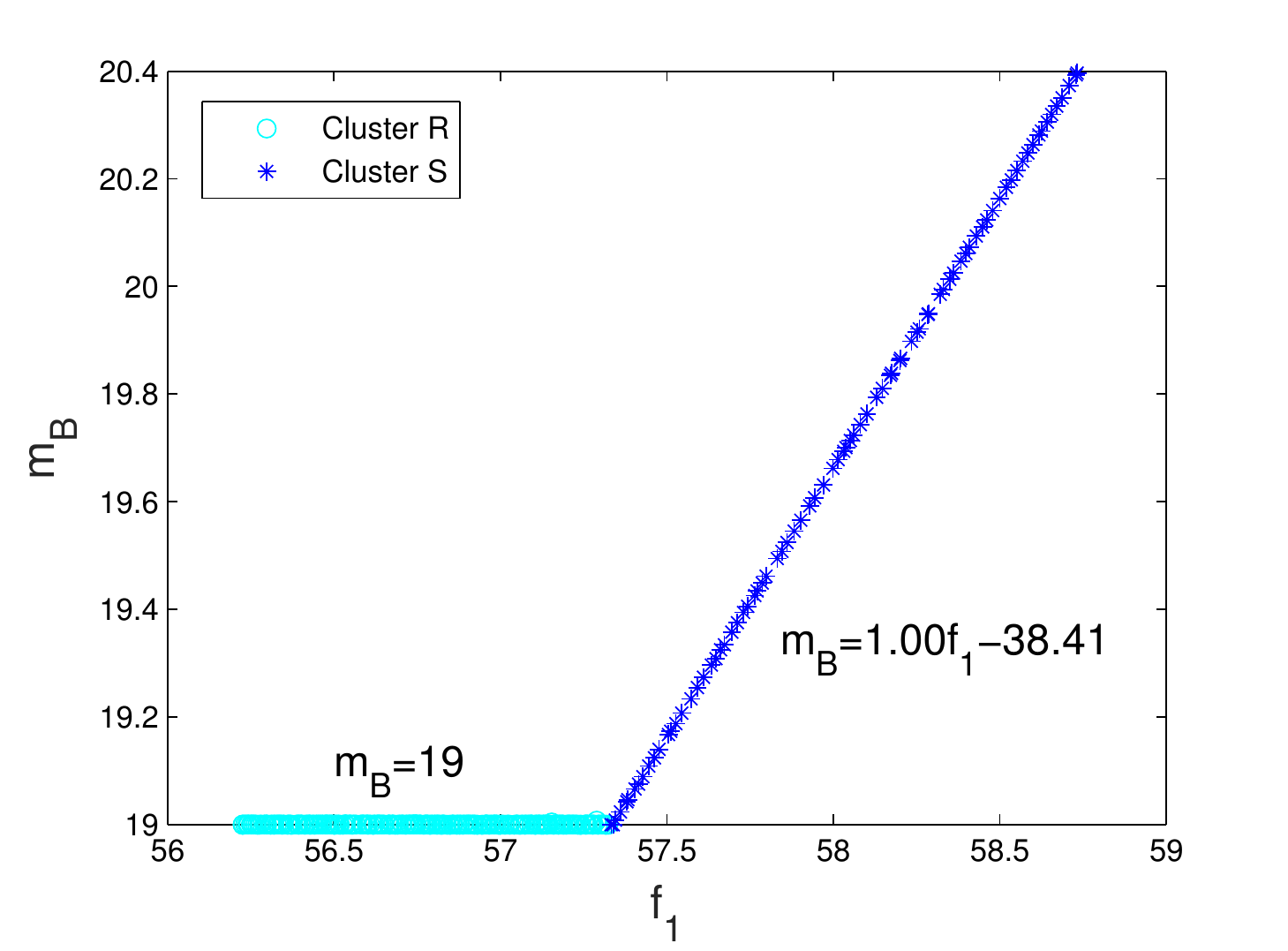}
\caption{Variation of the mass of Counterweight B $m_B$ with the total mass is shown. $m_B$ stays at the lower bound first, then rises up along with a straight line.}
 \label{mBrep}
\end{minipage}
\hspace{1.2cm}
\begin{minipage}[t]{0.45\linewidth}
\centering
\includegraphics [width=7.5cm]{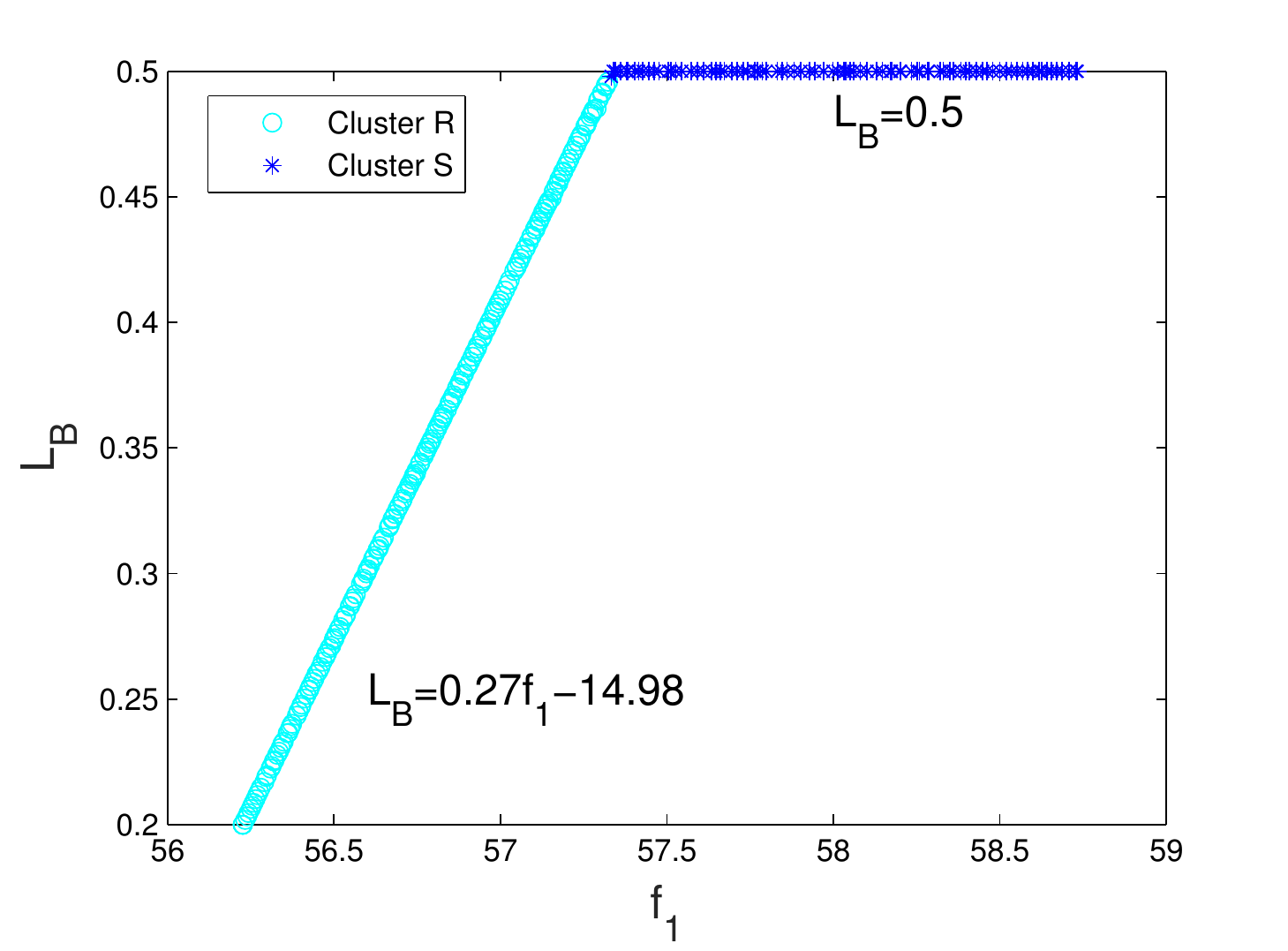}
\caption{Variation of the length of Counterweight B connecting rod $L_B$ with the total mass is shown. $L_B$ increases with a slope at 0.27 until reaching the upper bound, then keeps at 0.5. }
 \label{LBrep}
\end{minipage}
\end{tabular}

\begin{tabular}{cc}
\begin{minipage}[t]{0.45\linewidth}
\centering
\includegraphics [width=7.5cm]{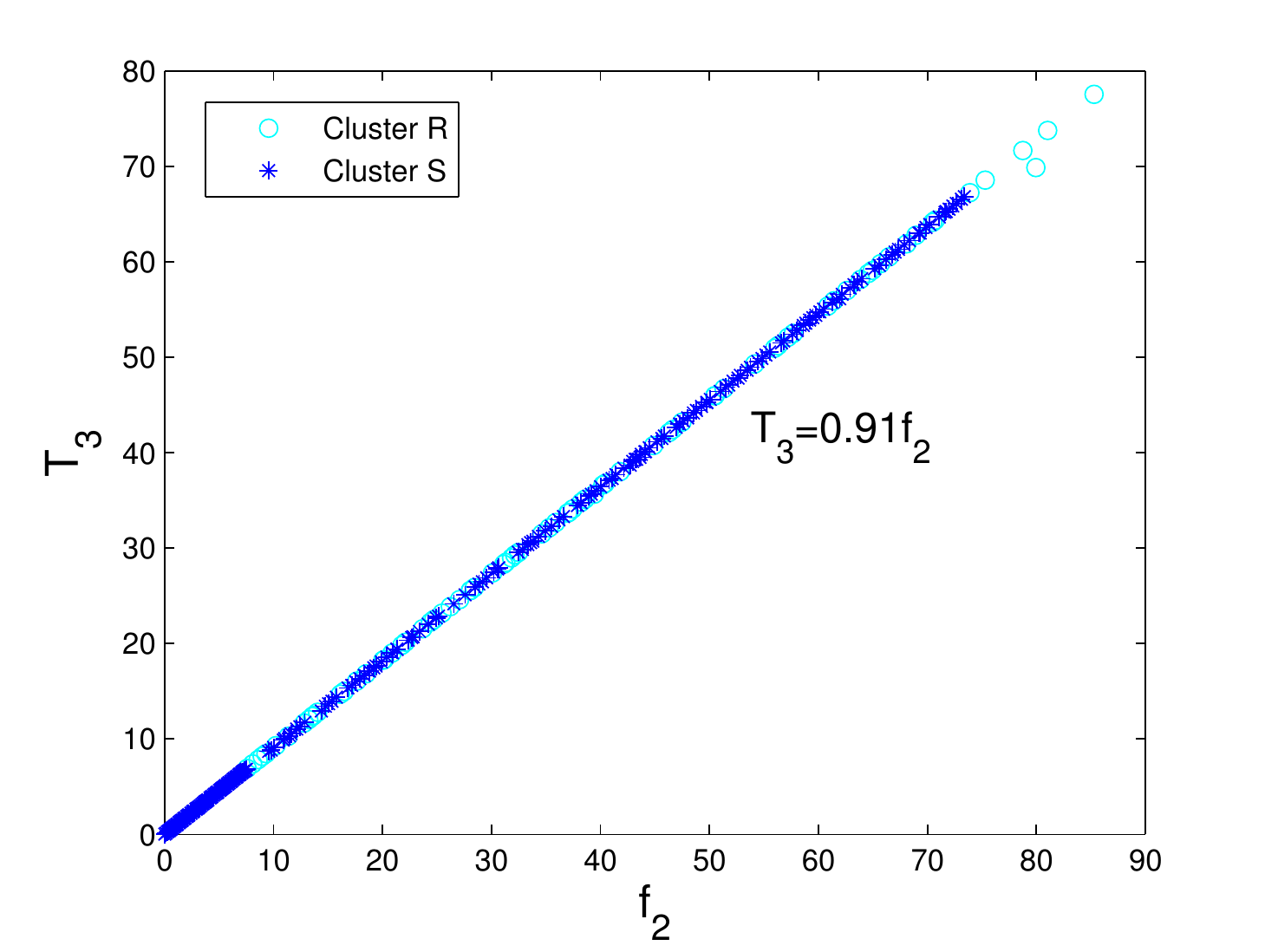}
\caption{Variation of the fiction moment of Fiction Disk 3 $T_3$ with the maximal needed operating force $f_2$ is shown, which is a straight line with a slope of 0.91. }
 \label{T3re2p}
\end{minipage}
\hspace{1.2cm}
\begin{minipage}[t]{0.45\linewidth}
\centering
\includegraphics [width=7.5cm]{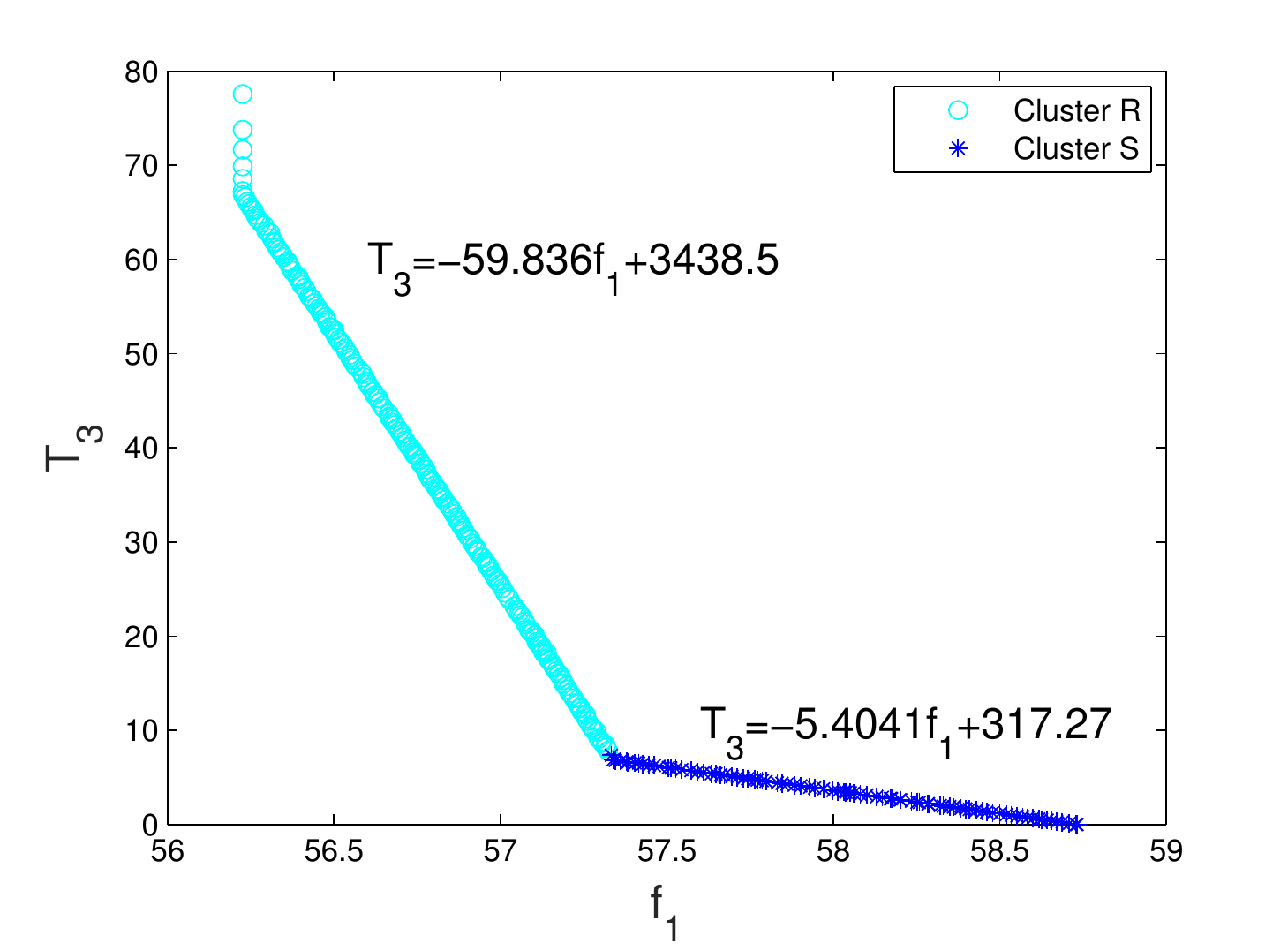}
\caption{Variation of the fiction moment of Fiction Disk 3 $T_3$ with the total mass $f_1$ is shown. $T_3$ decreases quickly in Cluster R and decreases slowly in Cluster S, with a turning point at 57.33.}
 \label{T3rep}
\end{minipage}
\end{tabular}
\end{figure*}

\begin{figure*}
\begin{tabular}{cc}
\begin{minipage}[t]{0.45\linewidth}
\centering
\includegraphics [width=7.5cm]{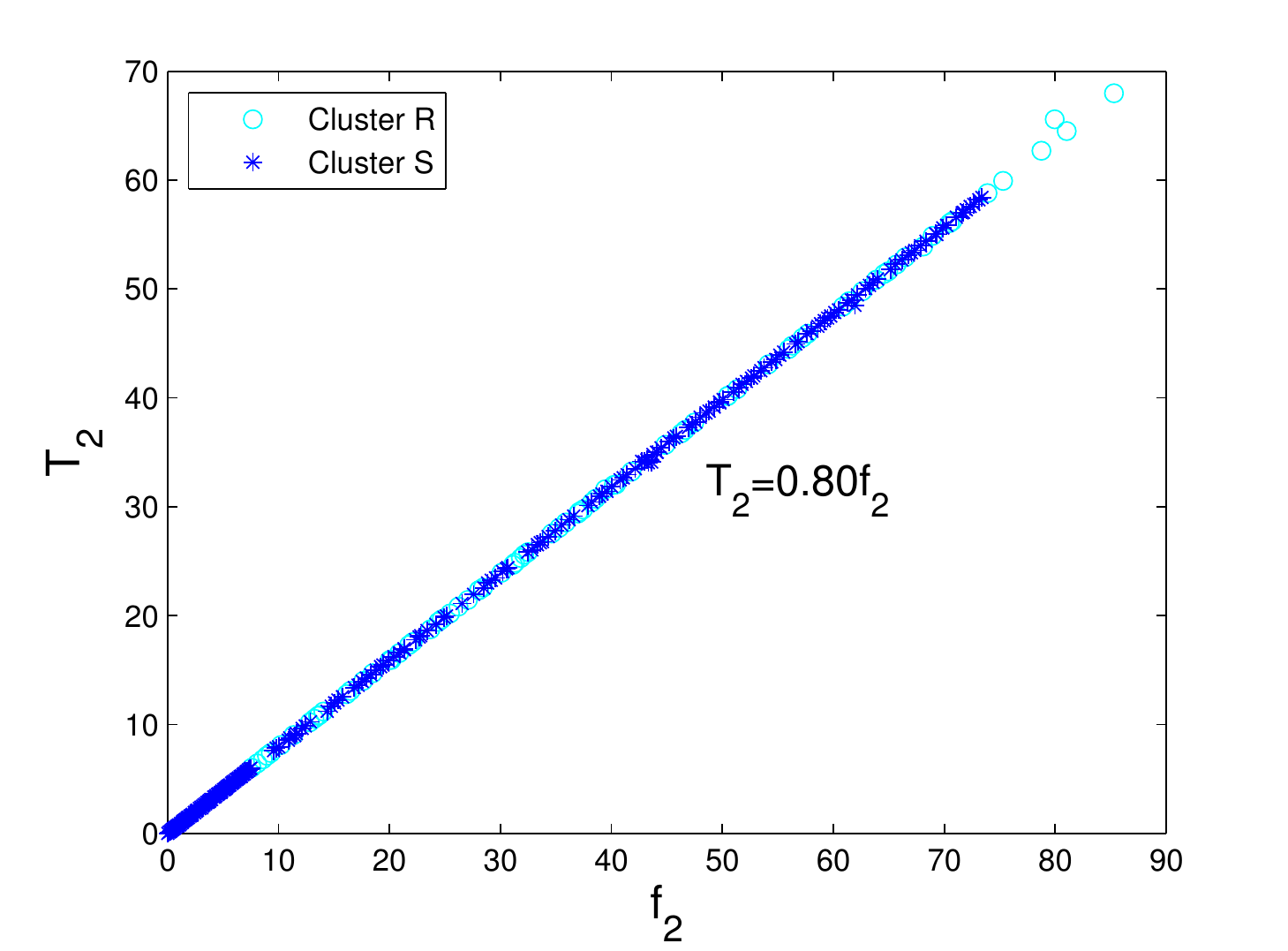}
\caption{Variation of the fiction moment of Fiction Disk 2 $T_2$ with the maximal needed operating force $f_2$ is shown, which is a staight line with a slope of 0.80.}
\label{T2re2p}
\end{minipage}
\hspace{1.2cm}
\begin{minipage}[t]{0.45\linewidth}
\centering
\includegraphics [width=7.5cm]{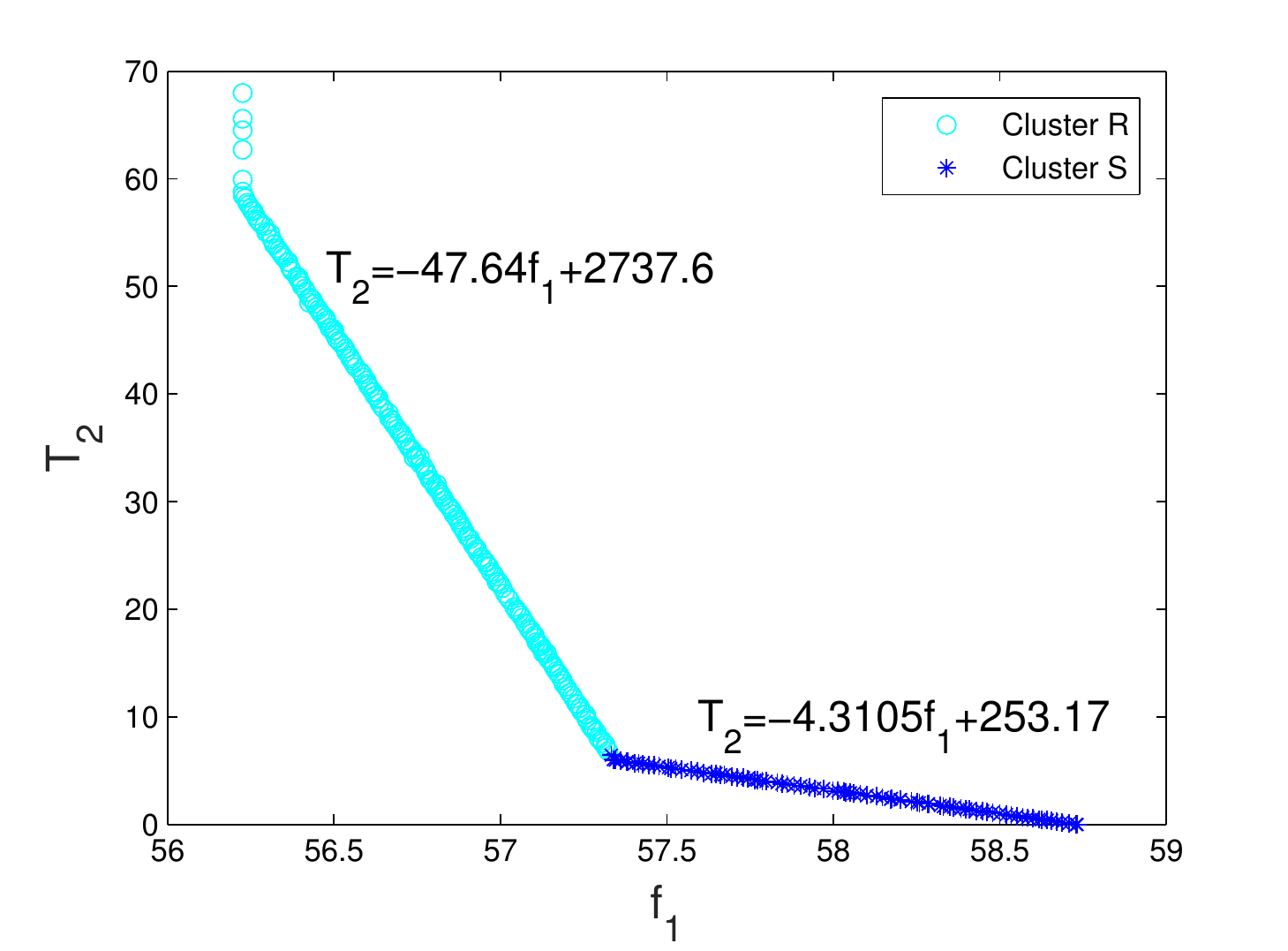}
\caption{Variation of the fiction moment of Fiction Disk 2 $T_2$ with the total mass $f_1$ is shown. $T_2$ decreases quickly in Cluster R and decreases slowly in Cluster S, with a turning point at 57.33.}
\label{T2rep}
\end{minipage}
\end{tabular}

\begin{tabular}{cc}
\begin{minipage}[c]{0.45\linewidth}
\centering
\includegraphics [width=8cm]{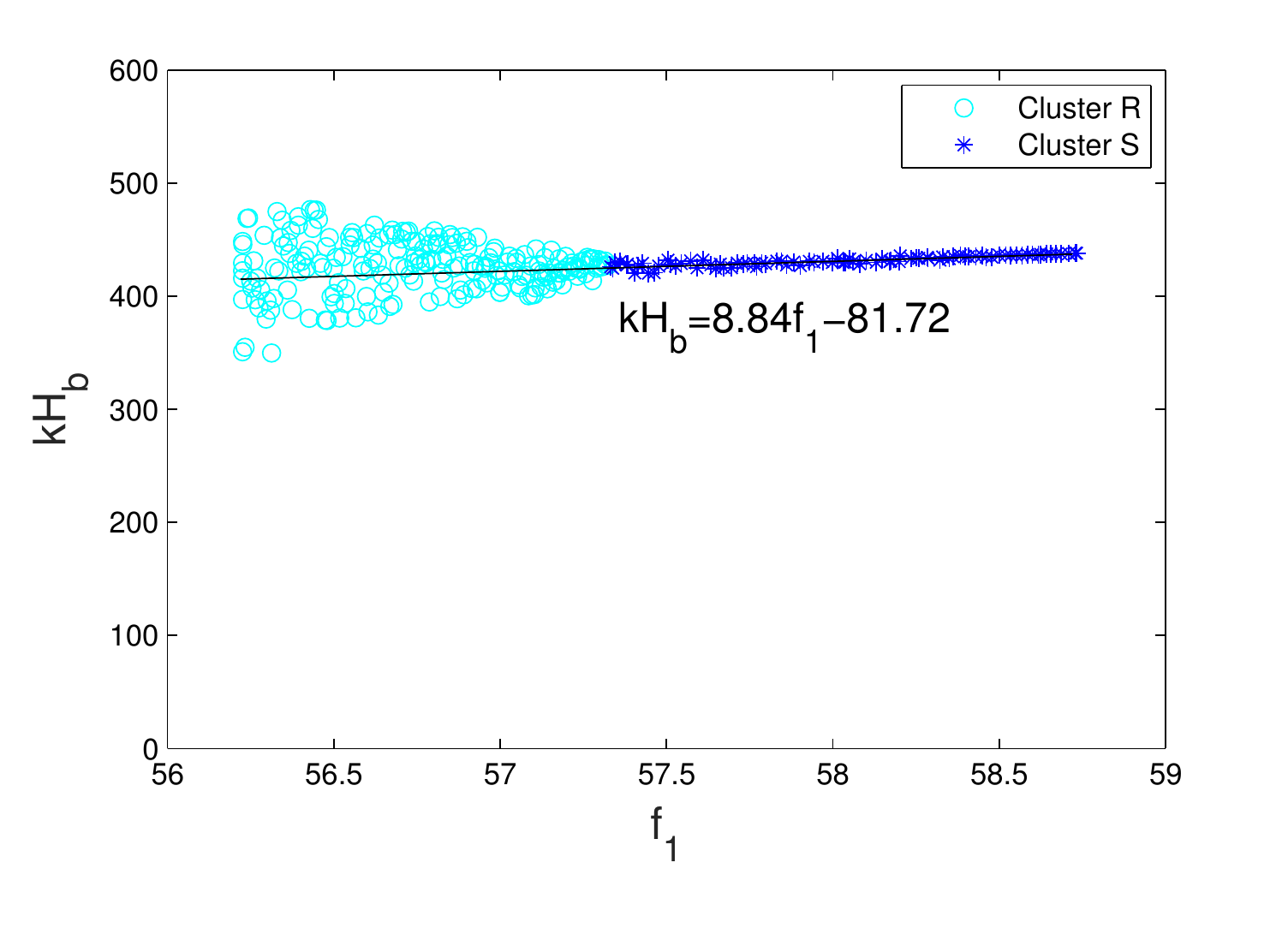}
\caption{Variation of product $kH_b$, which represents the tendency of the drafting force of the balancer, with the total mass $f_1$ is shown. The mean of $kH_b$ is about in a line. However, a specific $f_1$ can match different values of $kH_b$ in Cluster R. The range of $kH_b$ becomes smaller along with the increase of total mass until focusing at about $f_1=57.33$, then keeps increasing in a line.}
\label{KHbrep}
\end{minipage}
\hspace{1.2cm}
\begin{minipage}[c]{0.45\linewidth}
\centering
\includegraphics [width=8cm]{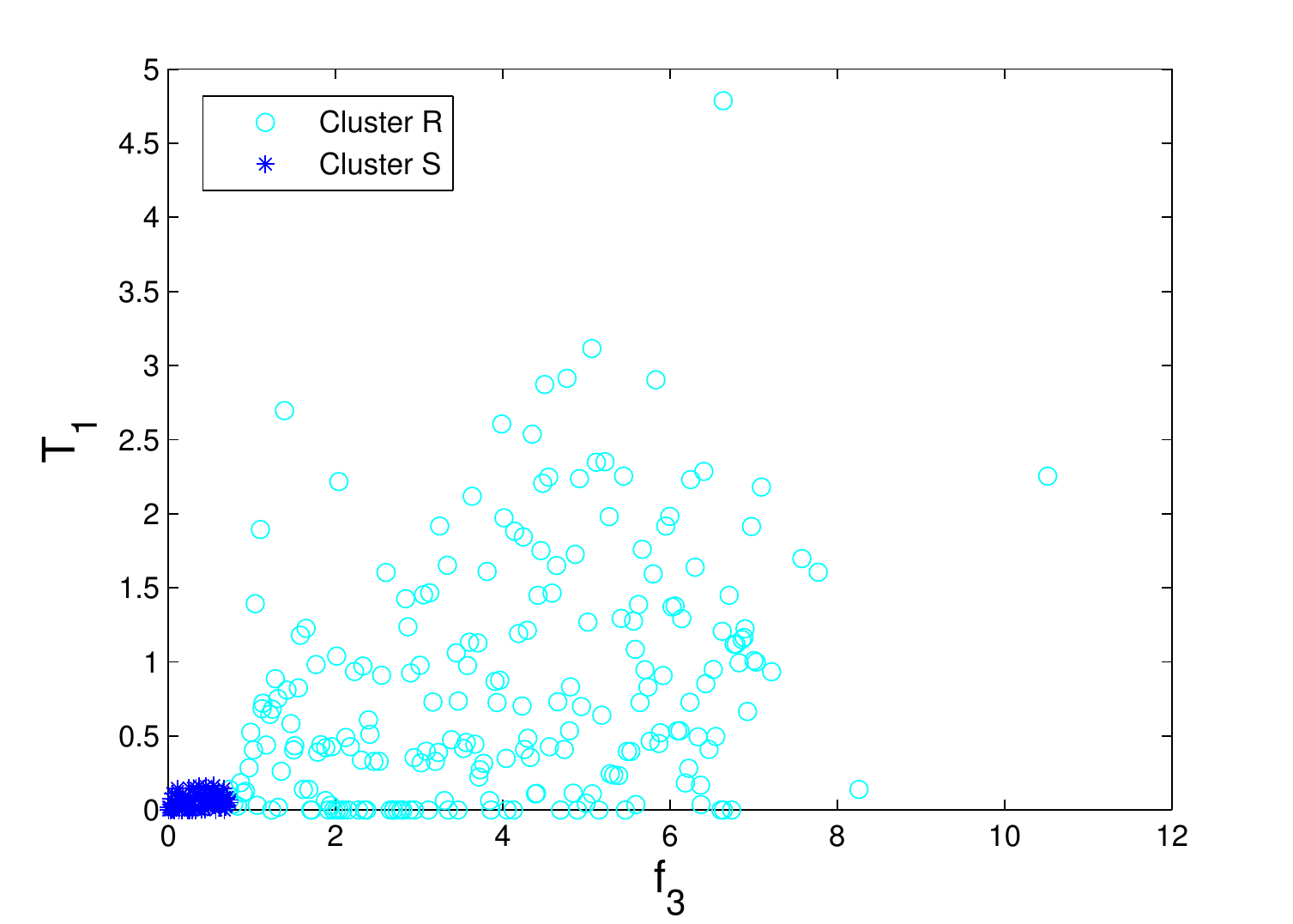}
\caption{Variation of the fiction moment of Fiction Disk 1 $T_1$ with the total mass $f_3$ is shown. There is a range between 0 to about 5.}
\label{T1re3p}
\end{minipage}
\end{tabular}
\end{figure*}

In this section, we perform an innovization study to discover some meaningful hidden relationships between objectives and design variables. There should be some common principles among all or part of the optimal solutions. Then the common principles can help the designer in future design, e.g. the knowledge discovered from the teaching manipulator design can be reused when the designer performs a similar teaching manipulator design again.
Fig. \ref{mArep} - \ref{T1re3p} illustrate the results of innovization study. As examples, we discuss the situation of the three gravity sensitive joints, including Joint 2, 3 and 5.

At Joint 5, it is easy to notice that $m_A$ is fixed at 1.44. while $L_A$ is fixed at 0.2, which is the upper bound of $L_A$. The couple of valuables keep Joint 5 balance with minimal mass and decrease the load of the other two joints. The relationships with $f_1$ are shown in Fig. \ref{mArep} and \ref{LArep}.

At Joint 3, Counterweight B, its connecting rod and a friction disk contribute in balancing. Fig. \ref{mBrep} shows the relation of $m_B$ vs $f_1$  and and Fig. \ref{LBrep} shows the relation of $L_B$ vs $f_1$. Both of the curve can be divided into two clusters (Cluster R and S), representing different situations. The two clusters are in different linear relations, with Eq.(\ref{mBre}) and (\ref{LBre}), respectively.

\begin{equation}
m_B=\left\{\begin{array}{ll}
19&f_1\in[56.22, 57.33],\\
1.00f_1-38.41&f_1\in[57.33, 58.72],
\end{array}\right.
\label{mBre}
\end{equation}
\begin{equation}
L_B=\left\{\begin{array}{ll}
0.27f_1-14.98&f_1\in[56.22, 57.33],\\
0.5&f_1\in[57.33, 58.72],
\end{array}\right.
\label{LBre}
\end{equation}

Total mass $f_1=57.33$ is an important turning point of the curve. In cluster R, $m_B$ keeps at 19, which is its lower bound, while $L_B$ linearly increases until reaching its upper bound, with a slope at 0.27 along with $f_1$ which is the reciprocal of the mass per unit length of the connecting rod B ($\rho_B=1/2.7=3.7$). In cluster S, $m_B$ have a steady increase with a slope at 1.00 along with $f_1$, while $L_B$ stay at 0.5. It means that an increase of unit length of connecting rod contributes less total mass than an increase of unit weight of Counterweight B. To keep minimal total weight, the optimal solution tend to increase $L_B$ first and to increase $m_B$ only when $L_B$ reach its upper bound.

In Fig. \ref{T3re2p}, it is illustrated that there is a linear relation between $T_3$ and $f_2$, with the equation
\begin{equation}
T_3=0.91f_2
\label{T3re}
\end{equation}

It is shown that there is a trade-off between the mass of counterweight B, the length of the connecting rod and the disk 3 fiction moment when comparing Fig. \ref{T3rep}, \ref{mBrep} and \ref{LArep}. The moment to rotate friction disk 3, namely $T_3$, decreases quickly in cluster R, along with the increase of total mass because the increase of $L_B$ can provide larger moment for balancing. $T_3$ decreases more slowly in cluster S, along with the increase of $m_B$.

At Joint 2, the gravity balance is kept by the trade-off of the balancer and friction disk 2. The situation of $T_2$ is similar to $T_3$ In Fig. \ref{T2rep}, the plot shows the relation between the $T_2$ vs $f_1$, where two clusters are in different decreasing lines with a turning point at $f_1=57.33$. In Fig. \ref{T2re2p}, the plot shows a linear relation between $T_2$ and $f_2$. The equation is given as follow.
\begin{equation}
T_2=0.80f2
\label{T2re}
\end{equation}

Based of Eq.(\ref{balancerForce}), as $g$ and $L_2$ are constant, when $|\cos{q_2}|=1$, $kH_b$ can represent the tendency of the drafting force of balancer. $kH_b$ vs $f_1$ is shown in Fig. \ref{KHbrep}. The variant of $kH_b$ become smaller and its mean increasing slightly, along with the increase of total mass because the increase of mass leads to a larger moment for balancing in need. To simplify the relation, we can treat them with a linear relation base on the mean value and ignore the variant of $kH_b$, with the equation as following.

\begin{equation}
kH_b=8.84f_1-81.27, f_1\in[56.22, 58.72],
\label{kHbre}
\end{equation}

It is noticed that the dots of cluster R in Fig. \ref{KHbrep} distributed in a  triangle region. From Eq.(\ref{me}) and (\ref{totomass}), $M_e$ is linearly proportioned to $f_1$. In cluster R, $T_2$ decreases rapidly along with the increase of $f_1$. In Eq.(\ref{balance2sim}), with a fixed $M_e$, the less $T_2$, the smaller range of $kH_b$ value satisfy the equation, namely a smaller range of the drafting force of the balancer. The variant of $kH_b$ represents the range of adjustment in terms of the value of the friction moment of friction disk 2 in Cluster R. As the total mass increases, the range of adjustment becomes smaller and gets into a line.

Fig. \ref{T1re3p} is the plot of $T_1$ vs $f_3$. The variation of $T_1$ is in a range from 0 to 5 $N\centerdot m$. As $f_3$ is in a small value, $T_3$ is nearly 0. From Fig. \ref{VSorg}, $f_1$ is small when the $f_3$ keeps in large value, because the solutions are tend to have lighter counterweights but mount the friction disks with larger friction moments for balance, which probabily lead to great variant of the operating force in the moving process. Therefore, the friction disks in Joint 1 is used for smoothing the resultant of the operating force in a trajectory.

Through the innovization study, we can establish some specific relations between the objectives and design variables. Meanwhile, we have deeper understanding about the interaction of the design variables of the optimal solutions. The above knowledge is difficult to be discovered in problem formulation or normal design procedure (e.g. the linear relation of $kH_b$ and $f_1$). With the knowledge, the designer can design a new teaching manipulator for other applications without a need to repeat solving the optimization problem again.

\section{Conclusion}
\label{sec:conc}
This paper focuses on modeling and optimizing a teaching manipulator. In the modeling stage, we formulate the balancing conditions of the three gravity sensitive joints and modeling of the operating force performance. The optimization stage shows the procedure to formulate and solve a three-objective constained design optimization problem. An innovization study is conducted to acquire a deeper understanding about the implicit design principles among multiple solutions.

The three objective functions include the minimization of the total mass of the device, the maximal operating force needed and the difference between the maximum and minimum of operating force. An evolutionary multi-objective optimization algorithm, NSGA-CDP is used to solve the multi-objective optimization problem. Compared with the original design of a human expert, the obtained solutions on the Pareto front are better in all the three objectives.

A comprehensive innovization study is conducted. The optimal solutions are used for data mining the implicit knowledge in the optimization problem. The relation equations between the objectives and design variables are established. Meanwhile, the interactions among objectives and the variables are discussed. The obtained knowledge can help the designer to make decisions more efficiently and effectively in a future design procedure.

As the next step, we will extend the research of innovization study. In the paper, we summarize the principles with visualization methods through human observation. However, many methods have been developed in data mining to automate the process of innovization \cite{deb2014integrated, bandaru2015generalized}. It will be of our great interest to apply these methods as more powerful tools to extract useful information from the design automation process.




\section*{Reference}
 \bibliographystyle{elsarticle-num}
 \bibliography{Myfile.bib}





\end{document}